\documentclass[lettersize,journal]{IEEEtran}
\usepackage{amsmath,amsfonts}
\usepackage{array}
\usepackage[caption=false,font=normalsize,labelfont=sf,textfont=sf]{subfig}
\usepackage{textcomp}
\usepackage{stfloats}
\usepackage{url}
\usepackage{verbatim}
\usepackage{graphicx}
\usepackage{cite}
\usepackage{multirow}
\usepackage{colortbl}
\usepackage{xcolor}
\usepackage{graphicx}
\usepackage{epstopdf}
\usepackage{amssymb}
\usepackage{algpseudocode}
\usepackage{algorithm}
\usepackage{booktabs}
\usepackage{multirow}
\usepackage{amstext}
\usepackage{amsmath}
\usepackage{csquotes}
\usepackage{epigraph}
\usepackage{amsmath,lipsum}
\usepackage{cuted}
\usepackage{amsfonts,amssymb}
\usepackage{enumitem}
\usepackage{mathrsfs}
\usepackage[section]{placeins}
\usepackage{float}
\usepackage{CJK}
\begin{document}

\title{Towards Improving Interpretability of Language Model Generation through a Structured Knowledge Discovery Approach}

\author{Shuqi Liu\textsuperscript{1}, Han Wu\textsuperscript{1}, Guanzhi Deng\textsuperscript{1}, Jianshu Chen\textsuperscript{3}, Xiaoyang Wang\textsuperscript{3}, Linqi Song\textsuperscript{1, 2, *\thanks{*Corresponding Author. This work was supported in part by the Research Grants Council of the Hong Kong SAR under Grant GRF 11217823 and Collaborative Research Fund C1042-23GF, the National Natural Science Foundation of China under Grant 62371411, InnoHK initiative, the Government of the HKSAR,Laboratory for AI-Powered Financial Technologies.}}\\
\textsuperscript{1}City University of Hong Kong\\
\textsuperscript{2}City University of Hong Kong Shenzhen Research Institute\\
\textsuperscript{3}Tencent AI Lab
}

\markboth{Journal of \LaTeX\ Class Files,~Vol.~14, No.~8, August~2021}%
{Shell \MakeLowercase{\textit{et al.}}: A Sample Article Using IEEEtran.cls for IEEE Journals}


\maketitle

\begin{abstract}
Knowledge-enhanced text generation aims to enhance the quality of generated text by utilizing internal or external knowledge sources. While language models have demonstrated impressive capabilities in generating coherent and fluent text, the lack of interpretability presents a substantial obstacle.
The limited interpretability of generated text significantly impacts its practical usability, particularly in knowledge-enhanced text generation tasks that necessitate reliability and explainability.
Existing methods often employ domain-specific knowledge retrievers that are tailored to specific data characteristics, limiting their generalizability to diverse data types and tasks. To overcome this limitation, we directly leverage the two-tier architecture of structured knowledge, consisting of high-level entities and lowlevel knowledge triples, to design our task-agnostic structured knowledge hunter. Specifically, we employ a local-global interaction scheme for structured knowledge representation learning and a hierarchical transformer-based pointer network as the backbone for selecting relevant knowledge triples and entities.
By combining the strong generative ability of language models with the high faithfulness of the knowledge hunter, our model achieves high interpretability, enabling users to comprehend the model’s output generation process. Furthermore, we empirically demonstrate the effectiveness of our model in both internal knowledge-enhanced table-to-text generation on the RotoWireFG dataset and external knowledge-enhanced dialogue response generation on the KdConv dataset. Our task-agnostic model outperforms state-of-the-art methods and corresponding language models, setting new standards on the benchmark.
\end{abstract}


\section{Introduction}

\IEEEPARstart{L}{anguage} models have proven to be highly effective in generating coherent and fluent text, such as GPT \cite{brown2020language} and PALM \cite{chowdhery2022palm}, resulting in exceptional performance in various text generation tasks. 
 {Their utilization has also been subject to criticism due to their inherent ``black box" nature, where the decision-making processes of the models are not transparent to human experts \cite{sun2022black, diao2022black}. This lack of interpretability can pose significant challenges, particularly in domains where model predictions need to be justifiable and explainable, such as in knowledge-intensive tasks. The ability to provide explanations for the model's decisions can increase transparency and accountability, improve trust, and enable better decision-making \cite{yin2022survey, mialon2023augmented}.} Therefore, the development of interpretable language models is becoming increasingly important, and several approaches are being explored to enhance the transparency of these models \cite{yasunaga2021qa, creswell2022selection}.

Recently, there has been significant progress in developing models that integrate language models with differentiable retrievers, such as REALM \cite{guu2020retrieval} and LAMBDA \cite{thoppilan2022lamda}, which have demonstrated impressive results in unstructured document retrieval. 
While for structured knowledge retrieval, building upon the success of language models, current state-of-the-art approaches are centered around the integration of structured knowledge into language models, including the generation of descriptive text over multiple knowledge triples \cite{xie2022unifiedskg}, the generation of question-answering over external knowledge bases \cite{liu2023pre}, the incorporation of auxiliary supervised tasks around structured knowledge \cite{ li2021improving}, reasoning knowledge paths
over knowledge graphs \cite{ ni2022hitkg}, and developing highly specialized models \cite{gong2019table, zhong2023knowledge}. However, these existing methods face challenges when dealing with general large-scale structured inputs, leading to limitations in interpretability performance, such as knowledge tracing and factual consistency. Furthermore, many of these models are restricted to specific domains, tasks, or datasets. Ideally, models should possess the capability to provide interpretability across various tasks and structured knowledge types, including internal and external structured
knowledge, tables, and knowledge graphs.

\begin{figure*}[tbp]
    \centering
    \includegraphics[width=12.5cm]{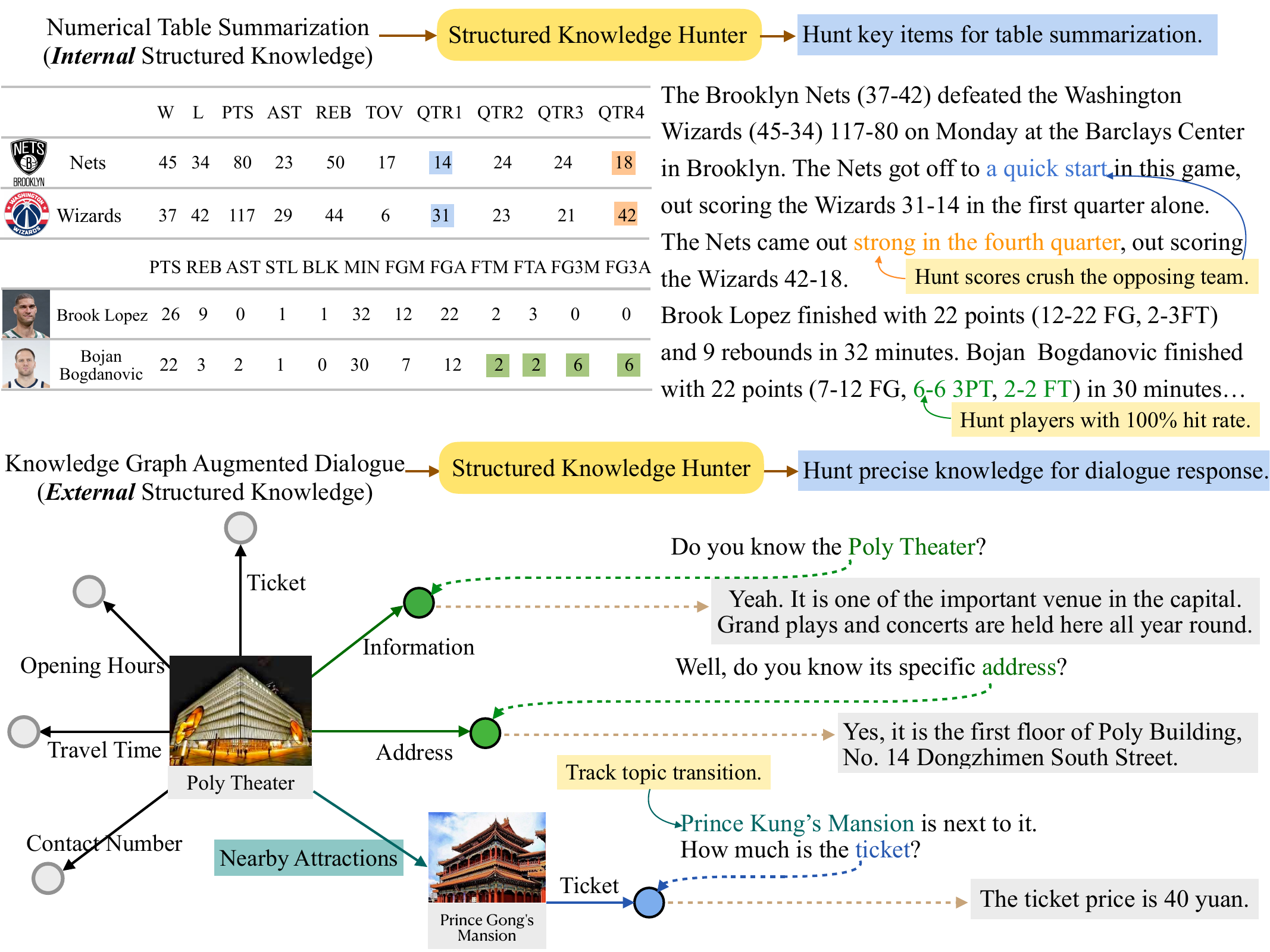}
    \caption{Generation examples of Structured Knowledge Hunter for numerical table summarization (top) and knowledge graph augmented dialogue (down). }
    \label{fig:intro}
\end{figure*}

In this paper, we propose a novel approach to augment language models with a structured knowledge retriever, which leverages the organized hierarchy of structured knowledge, such as tables, knowledge graphs, and databases, to store large-scale world or domain knowledge. 
To capture structured knowledge in a more interpretable way, we augment the language model with a differentiable structure knowledge retriever, which allows the model to retrieve and attend over tables or knowledge graphs from a large corpus (shown in Fig.~\ref{fig:intro}).
Specifically, we propose a task-agnostic Structured Knowledge Hunter to hunt the most relevant knowledge triples for language model-based text generators, thus leveraging redundant structured knowledge under the language model's capacity. 
To avoid the task and dataset-specific design, we fully leverage the two-tier architecture of structured knowledge with the high-level entity and low-level knowledge triples. 
Specifically, Structured Knowledge Hunter follows a local-global interaction manner whereby the structured knowledge representation is learned by a hierarchical encoder that fuses local intra-entity representation and global inter-entity representation, and the factual guidance extraction is learned by a Hierarchical Knowledge Planner through multi-task learning. Hierarchical Knowledge Planner is designed inspired by the ``macro-to-micro" selection behaviors of humans. ``Micro" knowledge triples are hunted under the influence of selected ``macro" entities, and multi-task learning encourages the matches between the hunted knowledge triple and the corresponding entity.

Our structured knowledge can be seen as a large external memory for language models to attend to. A key feature of our memory is that it is comprised of raw text rather than distributed representations, which makes the memory both (i) human-readable, lending a form of interpretability to our model, and (ii) human-writable, enabling us to dynamically update the model’s memory by editing the knowledge base.
The task-agnostic feature of our Structured Knowledge Hunter enables it to acquire generic factual guidance from structured input pertaining to various tasks and flexible types of structured knowledge (e.g., tables or knowledge graphs) through implicit learning.
We demonstrate this by both \textit{internal} structured knowledge-augmented Rotowire-FG table-to-text generation task \cite{wang2019revisiting} and \textit{external} knowledge graph-augmented KdConv dialogue response generation task \cite{zhou2020kdconv}. 
Evaluation results indicate that our pipeline paradigm outperforms state-of-the-art methods and the corresponding language models, thus setting new standards on the benchmark. Moreover, besides traditional text-level metrics, our Structured Knowledge Hunter also achieves higher knowledge selection accuracy relative to prior work with task-specific planning mechanisms.

Our contributions to this paper are as follows: 
\begin{itemize}
    \item  To the best of our knowledge, we are the first to directly use knowledge hierarchy to design a knowledge encoder, planner decoder, and optimization method for its task-agnostic ability to learn from large-scale structured input;
    \item 
    Our model combines the strong generation ability of language models with the high faithfulness of extractive knowledge selectors. As a result, our model can process generation tasks with high interpretability, allowing users to understand how the model generates its outputs. 
    
    \item We empirically demonstrate that our task-agnostic model can be adapted to both internal knowledge-enhanced table-to-text generation and external knowledge-enhanced dialogue response generation. It outperforms state-of-the-art methods and the corresponding language models, thus setting new standards on the benchmark.
\end{itemize}

\section{Related Work}
\paragraph{\textbf{Language Models: General-Purpose Architectures}} 
The previous research on general-purpose architectures for NLP tasks has demonstrated remarkable success, even without the use of retrieval. Specifically, a single pre-trained language model has proven to yield robust performance on a variety of classification tasks in the GLUE benchmark after fine-tuning \cite{wang2019superglue, devlin2018bert}. GPT-2 \cite{radford2019language} extended this approach by showing that a single, left-to-right pre-trained language model could excel in both discriminative and generative tasks. Furthermore, BART \cite{lewis2020bart} and T5 \cite{raffel2020exploring} propose a pre-trained encoder-decoder model that utilizes bi-directional attention to achieve even stronger performance on these tasks. In this paper, we aim to broaden the range of possible tasks by incorporating a retrieval module to enhance pre-trained generative language models' performance within a single, unified architecture.

\paragraph{\textbf{Languge Models in Knowledge-Augmented Text Generation}} 
Extensive research has been conducted on acquiring knowledge in information retrieval, particularly through the utilization of pre-trained neural language models (PLMs) \cite{nogueira2019passage, karpukhin-etal-2020-dense}, which greatly inspires our study.
One line is to directly insert the pre-computed knowledge representation as an auxiliary input to PLMs \cite{sun2019ernie, liu2021kg}. However, the method of explicitly injecting knowledge representation into PLMs has been argued that the vector space of PLM's words is inconsistent with knowledge representation since they are learned from separate tasks \cite{liu2020k}. 
Another line is to implicitly incorporate knowledge information into PLMs by performing knowledge-related tasks such as concept order recovering \cite{zhou2021pretraining} and entity category prediction \cite{yu2020jaket}. 
Others optimize the retrieval module to aid in a specific, downstream task using search \cite{menick2022teaching}, reinforcement learning \cite{liu2022rainier}, or a latent variable approach \cite{zhang2022subgraph}.
Moreover, RETRO \cite{borgeaud2022improving} extended by scaling the retrieval memory to trillions of tokens and changing the model architecture to take retrieved documents as input.
While previous successes have relied on various retrieval-based architectures and optimization techniques to achieve impressive performance on individual tasks, our study demonstrates that a single retrieval-based architecture achieves exceptional performance across a diverse range of tasks.

\paragraph{\textbf{\textbf{Table-to-Text Paragraph Generation}}} Unlike earlier table-to-text datasets, like WikiBio \cite{lebret2016neural} for sentence-level generation, RotoWire corpus \cite{wiseman2017challenges} is more challenging due to its paragraph-level target texts and more redundant contents in the table. 
The RotoWire corpus \cite{wiseman2017challenges} requires the model to generate NBA game summaries from the box- and line-score tables. Different from earlier table-to-text datasets, RotoWire is more challenging due to its longer target texts and more redundant contents in the table.
Some attempts find that content selection and planning are indispensable steps to good generation quality \cite{puduppully2021data, puduppully2022data}. To this end, they try to introduce auxiliary tasks to guide better content selection, and planning \cite{puduppully2019data,li2021improving, li2023plan}. However, these methods are not generalized due to task-specific objectives.
In contrast, Tableformer \cite{yang-etal-2022-tableformer} focuses on directly learning better table representations by modeling the table structure-aware features. 
Expressly, a hierarchical encoder \cite{gong2019table} represents the table by combining both the row- and column-level features to produce the target texts. 
The essential differences between our method and the previous models are 1) our model directly benefits from the knowledge hierarchy, thus being task-agnostic; 2) we use language models as text generators to produce more fluent and coherent texts.

\paragraph{\textbf{Knowledge Grounded Conversation}}
Since the standard dialogue response generation models
tend to produce dull and less informative responses, knowledge-grounded multi-turn dialogue generation \cite{zhao-etal-2022-standard, shuster-etal-2022-language} has attracted more attention recently. 
Knowledge-grounded conversation frequently uses a knowledge graph (KG) for the semantics in linked entities and relations \cite{wang2022rt, sarkar2022kg}. {KGs are also essential in fact verification to ensure the generation of faithful and accurate information \cite{yuan2023zero, aly2023qa}. }
KG-augmented dialogue generation encourages the model to generate diverse and meaningful content with the aid of retrieved knowledge. 
To address this task, one line of work learns KG-aware input text representations with graph neural networks \cite{xu2018graph2seq, kang2022knowledgeconsistent}. Another line performs reasoning over KGs via path-finding strategies to provide auxiliary guidance for the generation process \cite{moon2019opendialkg, tuan-etal-2022-towards}. Others explored fine-tuning the language models directly on the knowledge graph triple to transfer knowledge into a language model \cite{guan2020knowledge, yu2022jaket}. 
Instead, we use an explicit knowledge hunter and model more fine-grained features of the selected knowledge.

\section{Task Definition}



We define a general pipeline paradigm for different source knowledge-enhanced text generation tasks which can be customized for both internal and external knowledge-enhanced text generation tasks. 
We focus on developing a two-stage approach to generate coherent and faithful text using a structured knowledge input $K$. The input consists of $n$ knowledge triples and $p$ entities. The two-stage method is to first learn an extractive knowledge plan, denoted as $P_{m}$, which includes $m$ knowledge triples from $K$, and then be used by a language model-based text generator to produce high-quality text. 
To achieve this, we propose a generation process that involves selecting a knowledge triple $k_{t}$ at each $t$-th step based on a probability distribution $p(k_{t}|P_{t-1}, K; \theta)$, where $P_{t-1}$ represents the previously selected plan and $\theta$ denotes the model parameters. The selected triple is then added to $P_{t-1}$ to form $P_{t}$. 
We define the best plan $\hat{P}$ as the one that maximizes the probability of generating coherent and faithful text, given $K$ and $P_{m}$. 
\begin{equation}
    \hat{P} = \operatorname*{argmax}_{P_{m}, \forall m} \prod_{t=1}^{m} p(k_{t}|P_{t-1}, K; \theta)
\end{equation}

Suppose that we have a dataset $\mathcal{D}=\{X_{i}, K_{i}, y_{i}\}_{i=1}^N$, where $\forall i \in \{1, .., N\}$. 
In the context of dialogue response generation, the input $X_{i}$ represents the dialogue context, while $K_{i}$ is an external knowledge graph that contains relevant information associated with $X_{i}$. The objective is to generate a response $y_{i}$ based on $X_{i}$ and $K_{i}$. On the other hand, in the case of table-to-text generation, the input $X_{i}$ is a sequence of attribute-value tuples, and the output $y_{i}$ is a long descriptive text. In this case, $K_{i}$ is equivalent to $X_{i}$, as both represent the internal knowledge required to generate the output text.

\section{Model}

\begin{figure*}[htbp!]
\begin{center}
\includegraphics[width=0.95\textwidth]{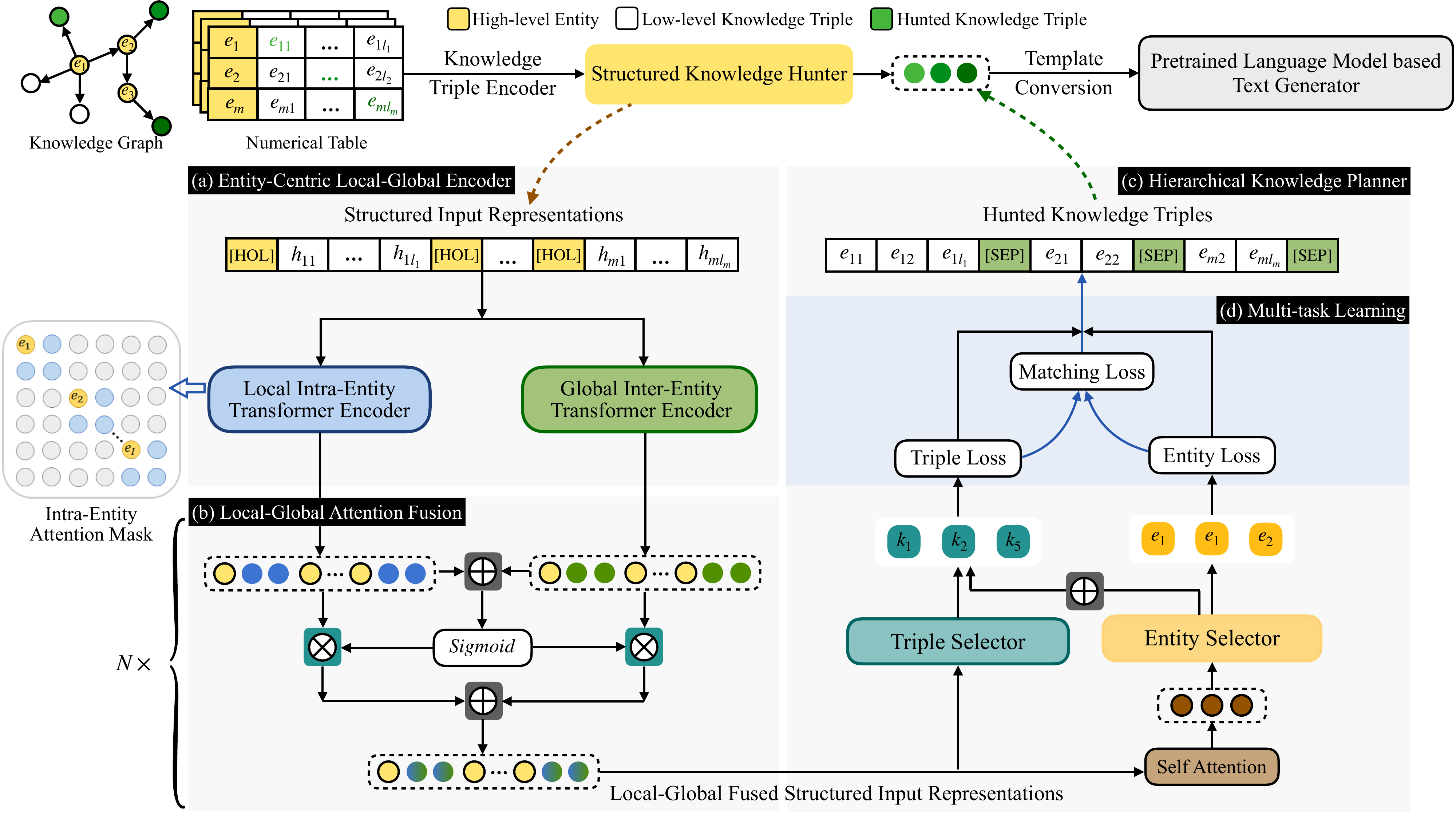}
\end{center}
\caption{Overall pipeline paradigm of steering language model-based text generator with a Structured Knowledge Hunter. The Knowledge Hunter is designed as a hierarchical encoder-decoder system that searches for relevant knowledge triples from input knowledge graphs or numerical tables. These knowledge triples are then processed through a Template Conversion stage that transforms them into a format that can be readily fed into the language model. Finally,
the language model generates text based on the processed knowledge triples. 
}
\label{model_arch}
\end{figure*}

The proposed pipeline paradigm for steering language model-based text generator with a Structured Knowledge Hunter is depicted in Figure \ref{model_arch}. In this paradigm, the Structured Knowledge Hunter module is responsible for processing structured knowledge and extracting associated knowledge triples that serve as factual guidance to the language model-based text generator. To make the extracted knowledge triples easily understandable, they are transformed into readable sentences using pre-defined templates. The transformed sentences are then used to fine-tune the language models to learn the factual knowledge guidance and improve text generation.

Specifically, (a)we introduce an Entity-Centric Local-Global Encoder that is designed to encode structured input representations, which have been embedded by a Knowledge Triple Encoder, in both local intra-entity and global inter-entity views. This allows us to capture both the details of individual entities as well as their relationships with other entities in the structured data;
(b) we utilize the Local-Global Attention Fusion module to aggregate the local-global scale context for fusion output. This module is crucial in ensuring that important information from both the local and global views is captured and effectively integrated into the final output.
(c) To generate high-quality knowledge triples, we introduce a Hierarchical Knowledge Planner that operates in a multi-task learning setting to jointly generate low-level hunted knowledge triples and corresponding high-level entities. This approach ensures the consistency of the hunted knowledge hierarchy and encourages matches between hunted knowledge triples and generated entities. By incorporating this hierarchical planner, we are able to generate more accurate and contextually relevant knowledge triples that can be used to enhance downstream tasks.
Overall, our approach combines several innovative techniques to improve the accuracy and reliability of structured knowledge representation. 
By leveraging both local and global views of entities, utilizing attention-based fusion, and incorporating a hierarchical knowledge planner, we are able to generate more accurate and contextually relevant knowledge triples that can be used to enhance various downstream tasks.


\subsection{Knowledge Triple Representation}


Knowledge triples are essential components of structured knowledge representations, including tables and knowledge graphs. Each triple in such representations captures either the relationships between different entities or the attribute-value pairs within a specific entity. 
For example, in a knowledge graph, a triple may represent a relationship between two entities, such as ``Person A works for company B." Alternatively, a triple may represent an attribute-value pair within a single entity, such as "Person A has the job title of the software engineer." In this way, knowledge triples serve as the building blocks of structured knowledge representations, allowing for the efficient storage and retrieval of information.

Thus, we formulate each knowledge triple as $<$\textit{entity name}, \textit{attribute name}, \textit{attribute value} $>$, where each attribute value could be a number or a string of text for diverse knowledge sources.

\paragraph{Numerical Table Embedding} For numerically rich tables, we follow previous work \cite{li2021improving} to learn triple representations from randomly initialized states. Considering each knowledge triple contains four embeddings: entity name $n_{ij}$, attribute name $a_{ij}$, attribute value $v_{ij}$, and attribute type $t_{ij}$. We encode each triple item into vector using concatenation $\rm [;]$ and 1-layer MLP with trainable parameters $W^{e}$ and $b^{e}$: 
\begin{equation}
    h_{ij}^{emb}=\text{Relu}(W^{e}[n_{ij};a_{ij};v_{ij};t_{ij}]+b^{e})
\end{equation}
where $i, j$ denotes the knowledge triple from $i_{th}$ entity and $j_{th}$ attribute. 

\paragraph{Context-Aware Knowledge Embedding} Otherwise, for text-rich knowledge tuples in dialogue response generation, we choose BERT \cite{kenton2019bert} as the backbone of our embedding module. 
Considering dialogue context $U= \{ u_{i} \}_{i=1}^{n}$ of $n$ utterances, where $u_{i}= (w_{1}^{i}, ..., w_{l_{u}}^{i})$ is a sequence of tokens of length $l_{u}$, 
and dialogue-related knowledge $K= \{ k_{i} \}_{i=1}^{m}$ of $m$ knowledge sentences from $p$ entities, where $k_{j}= (w_{1}^{j}, ..., w_{l_{k}}^{j})$ is a $l_{k}$ length token sequence, 
we define $\mathcal{S}=(S_{11}, ...S_{pm})$ as the input to BERT encoder. 
$S_{ij}$ concatenates the dialogue context and $j_{th}$ knowledge from $i_{th}$ entity:
   $ S_{ij} = {\rm[CLS]} w_{1},...,w_{l_{u}} {\rm[SEP]} w_{1}^{j},...,w_{l_{k}}^{j} {\rm[SEP]} $.
Thus, each context and knowledge pair could be sufficiently interacted through multi-layer attention in BERT, leading to context-aware knowledge representation. Based on the hidden state of the first special token {\rm [CLS]}, each knowledge representation is finally given by $e_{ij}^{emb}={\rm CLS} ({\rm BERT}(S_{ij}))$. 

\subsection{Entity-Centric Local-Global Encoder}



Based on the transformer architecture, the Entity-Centric Local-Global Encoder is a novel model for encoding knowledge triples in structured data. This encoder is comprised of two main components: a Local Intra-Entity Transformer Encoder and a Global Inter-Entity Transformer Encoder. The former component is designed to encode knowledge triples within the same entity, whereas the latter component is responsible for encoding structured data based on all underlying knowledge triples.

To enable the local and global encoding of knowledge triples, we employ different mask matrices $M$. These matrices play a crucial role in controlling what context a given knowledge triple can attend to when computing its contextualized representation. 
By using distinct mask matrices, we are able to tailor the encoding process to the specific needs of each component, thereby ensuring that the resulting representations are optimized for their respective tasks. Overall, by leveraging the Entity-Centric Local-Global Encoder and carefully designed mask matrices, we can achieve highly effective encoding of knowledge triples in structured data. 
The mask matrix for different item $i$ and $j$ is formulated as:

\begin{equation}
\label{mask_equ}
    M_{ij}=\left\{
\begin{aligned}
0 , \quad &  \ allow \ to \ attend \\
-\infty , \quad & otherwise \ no \ attending
\end{aligned}
\right.
\end{equation}

\paragraph{Local Intra-Entity Transformer Encoder} 
A local intra-entity transformer encoder enables the attention mechanism to focus only on the knowledge triples within the same entity, resulting in entity-aware knowledge representations. This is achieved through the use of a block diagonal matrix self-attention mask, which controls the attendance of each triple. The self-attention mask is composed of a main block diagonal part that is set to 0, and off-diagonal blocks that are set to $-\infty$ matrices. This is illustrated in Figure \ref{model_arch}(a), where the self-attention mask is referred to as Intra-Entity Attention Mask. Each square matrix $M_i$ in the mask corresponds to the attention mask for the $i_{th}$ entity, which may contain varying numbers of knowledge triples. 
The use of this self-attention mechanism in the local intra-entity transformer encoder allows for a more accurate and effective representation of entity-specific knowledge.

\paragraph{Global Inter-Entity Transformer Encoder} 
A global inter-entity transformer encoder enables all triples to attend to each other. In other words, the self-attention mechanism in this type of architecture allows for both inter-entity and intra-entity knowledge triples to be considered across all positions in the input sequence. This is achieved through the use of an all-zero matrix for the self-attention mask $M$, which effectively removes any limitations on the attention mechanism. As a result, the global inter-entity transformer encoder is able to capture more nuanced relationships and dependencies between entities, leading to improved performance in a variety of natural language processing tasks.

\paragraph{Attention Fusion Module} Attention Fusion Module aims to effectively combine the local and global contexts of a given input. 
This module accomplishes this task by allocating separate feature weights to each context, thereby allowing the model to selectively attend to relevant features. 
Given local view context $L$ and global view context $G$, the fusion output $H$ is aggregated as:
\begin{equation}
    H = \sigma(L\oplus G)\otimes L + (1-\sigma(L\oplus G)) \otimes G,
\end{equation}
where $\sigma$ is the Sigmoid function, $\oplus$ denotes the element-wise summation and $\otimes$ for the element-wise multiplication. 

Local weights $\sigma(L\oplus G)$ and global weights $1-\sigma(L\oplus G)$ are essential components of the Attention Fusion Module, which enable the network to perform a soft selection or weighted averaging between the local and global features. 
The local weights $\sigma(L\oplus G)$ and the global weights $1-\sigma(L\oplus G)$ are real numbers between 0 and 1, which control the degree of emphasis given to the local and global features during the fusion process. This mechanism allows the network to balance the importance of local and global information, resulting in a more accurate and robust learning model. 
Additionally, the Attention Fusion Module can be executed multiple times by iteratively using the output of the fusion process as the initial structured input representations. By doing so, the network can refine the learned representations and further enhance the performance of the model. This iterative process can be repeated up to $N$ times, where $N$ is a hyperparameter that can be tuned based on the specific task and dataset. 
Therefore, the Attention Fusion Module is a powerful technique for integrating local and global features in a flexible and effective way, leading to improved performance in structured data encoding.

\subsection{Hierarchical Pointer Planner}

We refer to the cognitive processes involved in planning sequential knowledge triples from structured inputs known as the “macro-to-micro” strategy. This strategy involves first identifying the entities mentioned in the input and then delving into each selected entity to determine the final required knowledge triples. This approach is frequently observed in human planning and reflects a top-down approach to information processing, where higher-level concepts are processed before lower-level details. By characterizing this selection behavior, our research sheds light on the underlying mechanisms of human cognitive processing and may have implications for the design of a knowledge planner.


With the Local-Global Fused Structured Input Representations $H$ as input, the Hierarchical Knowledge Planner generates a subset of relevant knowledge triples $K$ and their corresponding entities $E$. 
To handle the variability of the size of $K$ and $E$ across different contexts, we treat the construction process as a sequence generation task. At each step $t$, the planners add one knowledge triple or entity to $K$ or $E$, respectively, starting from an empty set and expanding it gradually. 
To ensure that each added knowledge triple or entity is selected directly from its structured input, we employ a transformer-based pointer network as the backbone of both the Triple Selector and the Entity Selector. 
This approach allows the model to effectively learn to select relevant knowledge triples and entities, enabling it to generate accurate and informative sequences of $K$ and $E$ for a given input $H$.

\paragraph{Entity Selector} 
To extract entity representations from $H$, we make use of the $\rm [HOL]$ special triple embeddings, which are gathered and passed through a self-attention module. This allows us to model entity-level embeddings $H_{e}=\{h_{e_{i}}\}_{i=1}^{p}$, where $p$ is the number of entities present in $H$.
To maintain a sequence of entity decoder hidden states $\{s_{t}^{e}\}_{t=0}^{T_{e}}$ and weight $\{e_{i}\}_{i=1}^{p}$, we employ an attention mechanism. Specifically, we compute the attention weights $e_{i}$ for each entity based on its relevance to the current decoder hidden state $s_{t}^{e}$. These weights are then used to compute a weighted sum of the entity embeddings $H_{e}$, which is subsequently incorporated into the decoder hidden state. 
This attention mechanism allows us to selectively focus on the most relevant entities at each decoding step, resulting in more accurate and effective entity representations.
\begin{equation}
P(e_{i}|e_{j_{1:t-1}}) = \frac{\exp (g(h_{e_{i}},s_{t}^{e}))}{\sum_{i}\exp (g(h_{e_{i}},s_{t}^{e}))},
\end{equation}
\begin{equation}
    g(h_{e_{i}},s_{t}^{e})=v_{e}^{\intercal} \tanh (W_{h_{e}}h_{e_{i}}+W_{e}s_{t}^{e}),
\end{equation}
where $W_{h_{e}}, W_{e}$ and $v_{e}$ are trainable parameters. These parameters are used to calculate the likelihood of the next entity being $e_i$ given the previous entities in the sequence. The index $j_t$ is then determined by taking the argmax of this distribution over all possible entities $i \in \{1, ..., p\}$, denoted in \eqref{eq:eqj}.
Once the index $j_t$ has been determined, the next entity $e_{j_t}$ is inserted into the sequence $E$.
\begin{equation}
    j_{t}={\rm argmax}_{i\in \{1, ..., p\}}P(e_{i}|e_{j_{1:t-1}})
    \label{eq:eqj}
\end{equation}

The decoder state $s_{t+1}^e$ is then calculated based on this new entity and the previous decoder state $s_t^e$. This calculation involves a combination of the entity embedding $H_e$, the previous decoder state $s_t^e$, and the output entity embedding from the attention mechanism $h_{j_{t}}$. 
\begin{equation}
    s_{t+1}^{e}={\rm TransformerDecoder}(h_{j_{t}}, s_{t}^{e}, H_{e})
\end{equation}

\paragraph{Knowledge Selector} Similarly, given structured knowledge representation $H=\{h_{i}\}_{i=1}^{m}$, we maintain a sequence of knowledge decoder hidden states $\{s_{t}^{k}\}_{t=0}^{T_{k}}$ and attention weight $\{k_{i}\}_{i=1}^{m}$ as

\begin{equation}
P(k_{i}|k_{j_{1:t-1}}) = \frac{\exp (g(h_{i},s_{t}^{k}))}{\sum_{i}\exp (g(h_{i},s_{t}^{k}))},
\end{equation}
\begin{equation}
    g(h_{i},s_{t}^{k})=v_{k}^{\intercal} \tanh (W_{h}h_{i}+W_{k}s_{t}^{k}),
\end{equation}
where $W_{h}, W_{k}$ and $v_{k}$ are trainable parameters. Next triple index $j_{t}$ is likewise selected by the max probability. Then element-wise summation is operated to combine entity and knowledge states for the calculation of $s_{t+1}^{k}$,
\begin{equation}
    s_{t+1}^{k}=s_{t+1}^{e} \oplus {\rm TransformerDecoder}(h_{j_{t}}, s_{t}^{k}, H).
\end{equation}

\subsection{Multi-task Learning} 

To ensure the accuracy and consistency of the knowledge hierarchy, we develop a hybrid optimization approach for our Structured Knowledge Hunter. Our optimization approach simultaneously optimizes three loss functions: the low-level knowledge triple loss $\mathcal{L}_{k}$, the high-level entity loss $\mathcal{L}_{e}$, and the matching loss $\mathcal{L}_{m}$. The matching loss is introduced to encourage the predicted knowledge triples to match with the predicted entities, thereby maintaining the hierarchical knowledge structure. 
Both $\mathcal{L}_{k}$ and $\mathcal{L}_{c}$ are computed using cross-entropy loss. Specifically, $\mathcal{L}_{k}$ measures the discrepancy between the predicted and ground-truth knowledge triples, while $\mathcal{L}_{e}$ measures the discrepancy between the predicted and ground-truth entities. 
This hybrid optimization approach ensures that our model learns to predict accurate knowledge triples while maintaining the hierarchical structure of the knowledge hierarchy.


\begin{equation}
    \mathcal{L}_{k}=-\sum_{t=1}^{|K|} \log P(k_{i}|k_{j_{1:t-1}}),
\end{equation}

\begin{equation}
    \mathcal{L}_{e}=-\sum_{t=1}^{|E|} \log P(e_{i}|e_{j_{1:t-1}}).
\end{equation}
while matching loss first maps the predicted knowledge triples into its corresponding entity by $map(\cdot)$ function, and then calculated the normalized distance between the mapped entity and the predicted entity by 
\begin{equation}
    \mathcal{L}_{m}=\frac{1}{p^{2}} \sum_{t=1}^{|K|} (map(K_{t}) - E_{t})^2
\end{equation}
The final loss we optimize is thus given by
\begin{equation}
    \mathcal{L} = \mathcal{L}_{k} + \mathcal{L}_{c} + \mathcal{L}_{m}
\end{equation}

\section{Experimental Setup}

In this section, we conduct experiments on two representing knowledge-enhanced text generation tasks to show the effectiveness of our method. We experiment with both internal knowledge-enhanced table-to-text generation on the RotoWire-FG dataset and external knowledge-enhanced dialogue response generation on the KdConv dataset. We describe our datasets, baselines, and implementation details here. 

\subsection{Datasets}

We experiment with representative datasets outlined below. 

\textbf{RotoWire-FG} \cite{wang2019revisiting} is a commonly used dataset for testing data-to-text generation models. It involves generating NBA game reports from box scores, which contain numerical tables capturing the statistical performance of both teams and individual players.
The dataset consists of 7476 pairs of box scores and summaries of NBA games from 2014 to 2019. The data is split into 5232 train, 1125 validation, and 1119 test examples. On average, there are 626 record triples in a box score per game. The average summary has 205.9 words and 8.6 sentences. 
By utilizing this dataset, we can evaluate our model's ability to accurately extract relevant details from numerical tables and generate concise summaries.

\textbf{KdConv} \cite{zhou2020kdconv} is a representative dataset for enhancing multi-turn dialogue response generation with the use of knowledge graphs. 
It is a multi-domain Chinese knowledge-driven conversation consisting of discussions on film, music, and travel. The dataset encompasses 4.5K conversations and 86K utterances, with an average of 19.0 turns and 2.3 topics per conversation. 
To utilize the pointer network-based knowledge planner effectively, we utilized the Longest Common Subsequence (LCS) algorithm to identify and label each attribute-value pair as an index within the dialogue knowledge. 
Through this dataset, we aim to demonstrate the effectiveness of our model in generating high-quality responses in complex, multi-domain, and knowledge-rich conversational contexts.

\subsection{Baselines}

For the RotoWire-FG dataset, we compare against the following six baselines.

\begin{itemize}[]
    \item Template \cite{wiseman2017challenges} method produces a summary by filling specified numbers into human-written templates. 
    \item CC \cite{wiseman2017challenges} is an LSTM-based encoder-decoder model with a conditional copy mechanism. 
    \item ENT \cite{puduppully-etal-2019-data} extends the CC model with an entity-specific dynamic memory module and generates a summary with a hierarchical attention mechanism. 
    \item NCP \cite{puduppully2019data} model employs a pointer network to select a subset of table records as content plan and then generates a summary using the CC model. 
    \item NCP + TR \cite{wang2019revisiting} extends the NCP model with a table reconstruction loss to improve fact grounding. 
    \item AuxEncoder \cite{li2021improving} is an entity-graph encoder with a reasoning module supervised by two auxiliary tasks of number ranking and importance ranking. 
\end{itemize}

For the KdConv dataset, we compare against the following baselines. 

\begin{itemize}
    \item Language Model (LM) \cite{NIPS2000_728f206c}: LM generates the next utterance through maximum log-likelihood estimation given a concatenated sequence of utterances from a dialogue. 

    \item Seq2Seq \cite{sutskever2014sequence} \& KSeq2Seq \cite{zhou2020kdconv}: Seq2Seq is an encoder-decoder model that incorporates an attention mechanism, allowing the model to focus on different parts of the input sequence while generating the output.  
    Kseq2Seq extends the Seq2Seq model by incorporating a key-value memory module to leverage knowledge information. All knowledge triples mentioned in a dialogue are treated as the knowledge information in the memory module. Key memory is the average word embeddings of the head entity and the relation, and the value memory is those of the tail entity. 

    \item HRED \cite{DBLP:conf/aaai/SerbanSBCP16} \& KHRED \cite{zhou2020kdconv}: HRED is a hierarchical recurrent encoder-decoder model. The past conversational utterances are integrated into a context state, which serves as the initial hidden state for the decoder. KHRED extends the HRED model by incorporating a key-value memory module, similar to Kseq2Seq.
\end{itemize}

\subsection{Implementation Details}

For the Rotowire-FG dataset, we adopt the settings from \cite{li2021improving} with a word embedding size of 600.
For the KdConv dataset, we follow the configurations outlined in \cite{zhou2020kdconv} by utilizing the Jieba Chinese word segmenter for tokenization \cite{sun2012jieba}. Each response includes two preceding history utterances, and the maximum knowledge length is limited to 256.
All transformer-based modules, including the intra-entity transformer encoder, inter-entity transformer encoder, topic transformer pointer, and knowledge transformer pointer, are composed of two transformer encoder or decoder layers with two multi-attention heads.
For the hierarchical knowledge planner, we employ the AdamW optimizer \cite{DBLP:conf/iclr/LoshchilovH19} with a learning rate of 1e-4. We set the dropout rate to 0.2 and use a beam size of 5 during inference. The knowledge planner is trained for 30 epochs, and the best checkpoint is selected based on the content selection F1 score on the validation set.
After obtaining the knowledge tuple indexes using the Structured Knowledge Hunter, we convert the selected knowledge tuples into textual representations using templates. The transferred knowledge plans are then realized through the text generator.
For the text generator, we employ the AdamW optimizer with a learning rate of 3e-5. During the decoding stage, for the RotoWire-FG dataset, we set the minimum decoding length to 20 and the maximum decoding length to 256. For the KdConv dataset, the maximum decoding length is set to 128. The text generator is fine-tuned for 20 epochs, and the best checkpoint is selected based on validation loss on the validation set.


\section{Results}


\begin{table*}[t]
\caption{The knowledge selection accuracy results for the RotoWire-FG dataset. These metrics include relation generation (RG) precision, content selection (CS) precision, recall, F1 score, and content ordering (CO) represented by the normalized Damerau-Levenshtein distance (DLD\%). Boldfaced numbers are the best results.}
    \centering
    \normalsize  
    \renewcommand\arraystretch{1.0}
    \setlength\tabcolsep{18pt}
    \resizebox{\textwidth}{!}{%
    \begin{tabular}{lccccc}
    \toprule
    \multirow{2}{*}{\textbf{Model}} & \multicolumn{1}{c}{\textbf{Relation Generation}} & \multicolumn{3}{c}{\textbf{Content Selection}} & \multicolumn{1}{c}{\textbf{Content Order}}  
      \\
    \cline{3-5}
        & \textbf{P} & \textbf{P} & \textbf{R} & \textbf{F1}& \textbf{DLD} \\ \hline 
    Template    & 93.72 & 23.98 & 43.96 & 31.03 & 10.25  \\ \hline
    CC \cite{wiseman2017challenges}   & 81.51 & 36.15 & 39.12 & 37.57 & 18.56  \\
    ENT \cite{puduppully-etal-2019-data}   & 98.89 & 39.04 & 49.29 & 43.57 & 17.5  \\ 
    NCP \cite{puduppully2019data}   & 94.21 & 43.31 & 55.15 & 48.52 & 23.46  \\ 
    NCP + TR \cite{wang2019revisiting}    & 95.70 & 42.90 & 56.91 & 48.92 & 24.47  \\ 
    AuxEncoder \cite{li2021improving}   & 94.75 & 42.72 & 57.56 & 49.04 & 25.23 \\ \hline
    T5  & 92.75 & 43.62 & 56.09 & 48.98 & 23.26  \\ 
    BART   & 94.71 & 44.58 & 57.24 & 50.12 & 24.35  \\ 
    HunterAug-T5  & 97.15 & 48.36 & 69.98 & 53.28 & 25.23   \\ 
    HunterAug-BART & \textbf{97.53} & \textbf{49.65}& \textbf{71.40} & \textbf{55.78} & \textbf{26.74}   \\ \bottomrule
    \end{tabular}
    }
    \label{tab:roto_table_main}
\end{table*}

\subsection{Table-to-Text generation}

We now describe the results on the RotoWire-FG dataset and compare our methods and the baselines in detail. In the table-to-text generation task, we conduct evaluations considering both knowledge selection accuracy and text quality. Knowledge selection accuracy is crucial for evaluating the interpretability of knowledge-enhanced text generation tasks. The extracted knowledge plan, which consists of relevant and accurate knowledge, forms the backbone of the generated text. This enables users to comprehend and trust the generated text, thereby improving overall interpretability. Additionally, we evaluate the text quality to ensure that the generated outputs are of high quality and coherent. 

\paragraph{\textbf{Knowledge Selection Accuracy}}
We employ the knowledge selection metrics \cite{wang2019revisiting}, such as Relation Generation (RG), Content Selection (CS), and Content Ordering (CO), to assess the model's ability to capture representative and relevant information from structured data. RG measures the precision of extracted triples from generated text compared to the box score table. CS considers the precision and recall of extracted triples from generated text compared to those from the target text. CO quantifies the normalized Damerau-Levenshtein distance between the triples extracted from the generated text and the target text.
To evaluate the accuracy and relevance of the knowledge included in the summaries, we train a knowledge extractor using ensemble models that combine convolutional and LSTM models. 
Table \ref{tab:roto_table_main} presents the evaluation results for knowledge selection accuracy. Initially, we assessed the generation and summarization capabilities of two state-of-the-art pre-trained language models (PLMs), T5 and BART. Subsequently, we integrated our Structured Knowledge Hunter into T5 and BART to enhance their ability to capture precise and comprehensive information during the text generation and summarization tasks. Among the tested PLMs, BART exhibited superior performance, surpassing AuxEncoder in content selection F1 score (50.12 F1\%). 

Following that, we proceeded to assess the performance of models augmented with the Structured Knowledge Hunter, specifically HunterAug-T5 and HunterAug-BART. The results demonstrate that our Structured Knowledge Hunter significantly improves the accuracy of hunted knowledge triples. 
The relation generation metric indicates that vanilla pre-trained language models like T5 and BART are less effective in utilizing factual knowledge compared to the previous state-of-the-art method, AuxEncoder. The integration of our Structured Knowledge Hunter into these models leads to significant improvements in relation generation performance. HunterAug-T5 achieves a relative increase of 2.53\% compared to AuxEncoder, while HunterAug-BART surpasses AuxEncoder with a relative increase of 2.93\%. 
Regarding content selection, vanilla BART already surpasses AuxEncoder in terms of F1 scores. When augmented with our knowledge hunter, it achieves a significant relative increase of 11.29\% compared to vanilla BART and a relative increase of 13.74\% compared to AuxEncoder. Similarly, equipping the T5 text generator with our knowledge hunter results in an 8.78\% relative increase compared to vanilla T5 and an 8.64\% relative increase compared to AuxEncoder in terms of content selection F1 score. 
In terms of content ordering, vanilla pre-trained models like T5 and BART exhibit lower performance compared to AuxEncoder. However, when enhanced with our Structured Knowledge Hunter, the content ordering results surpass the performance of AuxEncoder with a relative improvement of 5.98\%. The consistent improvement in knowledge selection accuracy suggests the effectiveness of our model in developing a better understanding of the relationships and dependencies among the information.

\begin{table}[htbp]
\caption{Text generation quality results for the RotoWire-FG dataset.}
\centering
\normalsize  
\renewcommand\arraystretch{1.0}
\setlength\tabcolsep{12pt}
\resizebox{0.48\textwidth}{!}{%
\begin{tabular}{lcccc}
\toprule
\multirow{2}{*}{\textbf{Model}} & \multirow{2}{*}{\textbf{BLEU}} & \multicolumn{3}{c}{\textbf{BERT Score}} \\
&       & P        & R        & F1   \\
\hline
AuxEncoder     &   24.52   &     76.37      &     73.17      &     75.07      \\
T5             &   23.34   &     77.83      &     73.90      &      75.81     \\
BART           &   24.51   &     79.17      &     73.63      &     76.30      \\
\hline
HunterAug-T5   &   26.87   &     \textbf{79.23}      &    76.18       &     77.68     \\
HunterAug-BART &   \textbf{28.69}   &     78.62      &      \textbf{78.12}     &       \textbf{78.37}      \\
\hline
\end{tabular}
}
\label{tab:roto_table_text_quality}
\end{table}

\paragraph{\textbf{Text generation Quality}} 
Text quality metrics assess the similarity and appropriateness of generated text compared to the ground truth. We employ automatic metric BLEU score, which measures n-gram overlap between generated and reference text. {We also utilize more recent metric BERTScore \cite{DBLP:conf/iclr/ZhangKWWA20}, which considers semantic similarity and contextual understanding for a comprehensive evaluation of text quality beyond simple n-gram matching.}
{According to the evaluation results in Table \ref{tab:roto_table_text_quality}, pretrained models enhanced with our structured knowledge hunter demonstrate superior performance in summary generation compared to the previous SOTA baseline model AuxEncoder, as well as pretrained language models T5 and BART. The enhanced models achieve impressive BLEU scores and BERT scores, indicating a significant overlap with the ground truth summary and accurate semantic representation of information. 
The BLEU score emphasizes the consistency of consecutive tokens, making it more sensitive to capturing factual information. On the contrary, BERTScore, being more semantic in nature, is less affected by factual information like numbers and names. Consequently, BERT scores tend to be higher when evaluating table summaries. 
Specifically, integrating our knowledge hunter model with pretrained models leads to an average BLEU improvement of 3.86 compared to vanilla pretrained models and a 3.26 improvement compared to AuxEncoder. Additionally, there is an average 1.97 BERTScore improvement compared to vanilla pretrained models and a 2.96 improvement compared to AuxEncoder.}

\begin{table*}[htbp]
    \caption{Automatic metrics for KdConv: BLEU score and distinct score. Boldfaced numbers are the best results.}
    \centering
    \normalsize
    \setlength\tabcolsep{13pt}
    \renewcommand\arraystretch{1.0}
    \begin{tabular}{lcccccccc}
    \hline
    \textbf{Model}
    & \multicolumn{4}{c}{\textbf{BLEU-1/2/3/4}} 
    & \multicolumn{4}{c}{\textbf{Distinct-1/2/3/4}}\\

    \hline 
    \multicolumn{9}{c}{\textbf{Travel}} \\ \hline
    LM \cite{bengio2000neural}  &  27.51 & 17.79 & 12.85 & 9.86 & 3.18 & 8.49 & 13.99 & 19.91  \\
    Seq2Seq \cite{sutskever2014sequence}  &  29.61 & 20.04 & 14.91 & 11.74 & 3.75 & 11.15 & 19.01 & 27.16  \\ 
    HRED \cite{serban2016building}  &  30.92 & 20.97 & 15.61 & 12.30 & 4.15 &  12.01 & 27.16 & 28.74 \\
    KSeq2Seq \cite{zhou2020kdconv}  &  37.04 & 27.28 & 22.16 & 18.94 & 4.25 & 13.64 & 24.18 & 34.08  \\ 
    KHRED \cite{zhou2020kdconv}  &  36.87 & 26.68 & 21.31 & 17.96 & 3.98 & 13.31 & 24.06 & 34.35 \\ 
    \hline
    HunterAug-BART &  \textbf{43.16} & \textbf{35.54} & \textbf{31.61} & \textbf{28.45} & \textbf{7.14} & \textbf{20.69} & \textbf{30.10} & \textbf{37.32}  \\ \hline
    
    \multicolumn{9}{c}{\textbf{Music}} \\ \hline
    LM \cite{bengio2000neural}  &  25.80 & 13.93 & 8.61 & 5.57 & 2.72 & 7.31 & 12.69 & 18.64  \\
    Seq2Seq \cite{sutskever2014sequence}    & 28.89 & 16.56 & 10.63 & 7.13 & 2.52 & 7.02 & 12.69 & 18.78 \\ 
    HRED \cite{serban2016building}    & \textbf{29.92} & 17.31 & 11.17 & 7.52 & 2.71 & 7.71 & 14.07 & 20.97 \\ 
    KSeq2Seq \cite{zhou2020kdconv}    & 29.60 & 17.26 & 11.36 & 7.84 & 3.93 & 12.35 & 23.01 & 34.23 \\ 
    KHRED \cite{zhou2020kdconv}   & 29.73 & 17.51 & 11.59 & 8.04 & 3.80 & 11.70 & 22.00 & 33.37 \\ \hline
    HunterAug-BART &  28.63 & \textbf{19.97} & \textbf{15.93} & \textbf{13.04} & \textbf{4.69} & \textbf{15.39} & \textbf{25.55} & \textbf{34.01}  \\ \hline
    
    \multicolumn{9}{c}{\textbf{Film}} \\ \hline
    LM \cite{bengio2000neural}  &  24.22 & 12.40 & 7.71 & 4.27 & 2.32 & 6.13 & 10.88 & 16.14  \\
    Seq2Seq \cite{sutskever2014sequence}    & 26.97 & 14.31 & 8.53 & 5.30 & 2.51 & 7.14 & 13.62 & 21.02 \\ 
    HRED \cite{serban2016building}    & 27.03 & 14.07 & 8.30 & 5.07 & 2.55 & 7.35 & 14.12 & 21.86 \\ 
    KSeq2Seq \cite{zhou2020kdconv}  & 27.45 & 14.51 & 8.66 & 5.32 & 2.85 & 7.98 & 15.09 & 23.17 \\ 
    KHRED \cite{zhou2020kdconv} & \textbf{27.94} & \textbf{14.69} & 8.73 & 5.40 & 2.86 & 8.08 & 15.81 & 24.93 \\ \hline

    HunterAug-BART & 22.29 & 14.58 & \textbf{10.96} & \textbf{8.50} & \textbf{5.57} & \textbf{16.26} & \textbf{27.38} & \textbf{37.63}  \\ \hline

    \end{tabular}
    \label{tab:kdconv_tab}
\end{table*}

\subsection{Multi-turn Dialogue Response generation}
We extend our model to handle text-rich, knowledge graph-driven tasks, specifically focusing on KdConv multi-turn dialogue response generation. We compare our methods with baselines, evaluating their performance in terms of knowledge selection accuracy and text quality.
For evaluating KdConv multi-turn dialogue response generation, we use BLEU score and Distinct score as specific metrics. BLEU score measures the overlap between generated and reference text based on matching n-grams. Distinct score quantifies the diversity of the generated text by calculating the number of unique n-grams present. These metrics provide insights into the quality and diversity of the generated responses.

We present text generation quality results on the KdConv Dataset in Table \ref{tab:kdconv_tab}. 
The Chinese BART model \cite{shao2021cpt} is utilized as the text generator for evaluating generation quality. 
Based on the results, the HunterAug-BART model demonstrates superior performance compared to all other models in terms of BLEU-k and distinct-k metrics. Specifically, when compared to the state-of-the-art baseline model KHRED, HunterAug-BART achieves significant improvements in BLEU-4 and Distinct-4 scores in the travel, music, and film domains. The BLEU-4 score is raised by 10.49/5.00/3.10, and the Distinct-4 score is raised by 2.97/0.64/12.70, respectively.
The increase in the distinct score highlights the enhanced generation capability of the pre-trained model utilized in HunterAug-BART. Moreover, the improvement in the BLEU score indicates the effective knowledge extraction ability of the hierarchical knowledge planner in selecting relevant knowledge for generating dialogue responses, as demonstrated by the knowledge selection score presented in Table \ref{tab:kdconv_ks}. 
The Knowledge Selection (KS) score evaluates the factual consistency of the generated text by assessing the precision and recall of selected knowledge tuples from a knowledge graph compared to those in the target response. KS quantifies the alignment between the generated text and the underlying knowledge graph, which is crucial for maintaining factual consistency in KdConv systems.

\begin{table}[ht]
    \normalsize
    \setlength\tabcolsep{14pt}
    \renewcommand\arraystretch{1.0}
    \caption{Knowledge Selection (KS) precision (P\%) and recall (R\%) for different domains in KdConv dataset.}
    \centering
    \begin{tabular}{lccc}
    \hline
    \textbf{Domain}
    & \textbf{KS-P\%}
    & \textbf{KS-R\%}
    & \textbf{KS-F1\%} \\
    \hline
    Travel & 88.00 & 86.54 & 87.13 \\ 
    Music & 85.48 & 82.39 & 83.65 \\ 
    Film & 80.72 & 74.60 & 76.95 \\ \hline
    \end{tabular}
    \label{tab:kdconv_ks}
\end{table}

The difference between the BLEU score and the knowledge selection score can be explained by the limitations of the KdConv dataset. The dataset only provides labeled responses that are relevant to knowledge, allowing the knowledge planner to accurately predict the absence of knowledge. However, the BART text generator may struggle to initiate a conversation when faced with incomplete or empty knowledge, which can impact the quality of the generated response and subsequently affect the BLEU score.
Our study also observed significant performance variations across different domains in terms of BLEU score and knowledge selection score. These variations can be attributed to statistical differences in the domain datasets. The film dataset presented greater challenges due to its higher knowledge diversity, consisting of 1837 entities and 318 relations, as well as a higher average utterance count per dialogue (24.4), compared to the travel dataset, which had a more limited knowledge base of 699 entities and seven relations, and an average of 16.1 utterances per dialogue.


\subsection{Ablation Study}
To validate the effectiveness of our Structured Knowledge Hunter, we conducted ablation studies to assess the impact of different model components on overall performance. Specifically, we focused on evaluating the hierarchical knowledge encoder (Table \ref{tab:encoder_ablation}) and the combination of loss functions (Table \ref{tab:loss_ablation}). 
We focus on the Content Selection score as it is crucial for understanding the model's decision-making process, explaining its performance and results, and guiding model improvements. It provides insights into how the model selects relevant information from structured knowledge and helps analyze its behavior during the knowledge extraction and selection process. 

\begin{table}[htbp]
\caption{Ablation study on hierarchical encoder}
\centering
\setlength\tabcolsep{8pt}
\resizebox{\columnwidth}{!}{%
\begin{tabular}{llll}
\hline
\multirow{2}{*}{\textbf{Encoder}} & \multicolumn{3}{c}{\textbf{Content Selection}}             \\
& \textbf{P\%} & \textbf{R\%} & \textbf{F1\%} \\
\hline
Plain Transformer Encoder         & 43.68 & 61.41  & 51.11  \\
+ Entity-Aware                    & 45.38 & 62.95  & 52.71  \\
+ Entity-Aware + 1 LG fusion      & 46.53 & 64.10  & 54.07  \\
+ Entity-Aware + 2 LG fusion      & \textbf{46.88} & \textbf{65.07}  & \textbf{54.49} \\
\hline
\end{tabular}%
}
\label{tab:encoder_ablation}
\end{table}

In the ablation study (Table \ref{tab:encoder_ablation}), we evaluated the effectiveness of the hierarchical encoder by gradually incorporating the entity-aware design and local-global fusion modules into a plain transformer encoder optimized solely using knowledge loss. The results showed that the plain transformer encoder combined with a transformer pointer decoder achieved satisfactory content selection results with an F1 score of 51.11\%. Introducing the entity-aware module, which considers structured knowledge hierarchically in an entity-centric manner, improved the F1 score by 1.60\%. This highlights the importance of modeling structured knowledge through an entity-centric approach for performance enhancement.
Additionally, we assessed the effectiveness of the local-global attention fusion (LG fusion) module by incorporating one and two LG fusion layers. The results demonstrated a 1.36\% improvement in the F1 score with the addition of one LG fusion layer. However, introducing an additional LG fusion layer did not yield a significant performance gain, suggesting that a single LG fusion layer is sufficient for optimal performance. 

\begin{table}[htbp]
\caption{Ablation study on loss functions.}
\centering
\normalsize  
\setlength\tabcolsep{8pt}
\renewcommand\arraystretch{1.0}
\resizebox{\columnwidth}{!}{%
\begin{tabular}{llll}
\hline
\multirow{2}{*}{\textbf{Loss}} & \multicolumn{3}{c}{\textbf{Content Selection}}             \\
                               & \textbf{P\%} & \textbf{R\%} & \textbf{F1\%} \\
\hline
Only Triple Loss     & 46.88 & 65.07 & 54.49  \\
+ Entity Loss            & 49.22 & 68.24 & 57.03   \\
+ Entity Loss + Matching Loss   & \textbf{50.16} & \textbf{71.61} & \textbf{57.25} \\     \hline
\end{tabular}%
}
\label{tab:loss_ablation}
\end{table}

Table \ref{tab:loss_ablation} reports the effectiveness of our multi-task learning approach using different loss combinations in the proposed Structured Knowledge Hunter architecture. Our experiments revealed that maintaining consistency within the knowledge hierarchy was crucial for accurate prediction of triples belonging to the correct entity. When using only the triple loss, we achieved an F1 score of 51.49\%, which improved by 2.54\% when the entity loss was added. {Furthermore, conducting additional tests with different random seeds showed a mean F1 score of 57.25 with a variance of 0.13 when using the matching loss. These results indicate that while the increase in F1 score may be marginal, the inclusion of the matching loss provides additional guidance and constraints that contribute to the overall performance of the model in content selection.} These findings offer valuable insights into the effectiveness of our proposed architecture and the impact of different loss function combinations on the model's overall performance.

\subsection{Human Evaluation}
{To assess the agreement between human judgments and improvements in automatic evaluation metrics, we conducted a human evaluation. Three proficient English-speaking graduate students were recruited for this task. Specifically, we randomly selected 30 games from the RotoWire-FG test set. Each game was independently rated by three evaluators on a scale of 0 to 5 across four criteria: Coherence (do the sentences, in summary, follow a coherent discourse?), Fluency (is the summary fluent and grammatical?), Information Coverage (does the summary covers more supported facts?), and Information Accuracy (does the information in the summary is factual and correct?). The human evaluation result is presented in Table \ref{tab:roto_table_human_eval}.}

\begin{table}[htbp]
\caption{Human Evaluation on the RotoWire-FG dataset.}
\centering
\normalsize  
\renewcommand\arraystretch{1.0}
\setlength\tabcolsep{3pt}
\resizebox{0.48\textwidth}{!}{%
\begin{tabular}{lcccc}
\toprule
\textbf{Model} & \textbf{Coherence} & \textbf{Fluency} & \textbf{InfoCov}        & \textbf{IncoAcc}   \\
\hline
AuxEncoder     &   2.97   &     3.67     &     3.30      &    2.93       \\
T5             &   3.20   &     4.20      &    3.47      &    3.17       \\
BART           &   \textbf{3.63}   &     \textbf{4.27}      &     3.87      &    3.37       \\
\hline
HunterAug-T5   &    3.40  &     3.57      &    4.10       &     3.97     \\
HunterAug-BART &    3.57  &     3.73      &    \textbf{4.13}       &    \textbf{4.03}      \\
\hline
\end{tabular}
}
\label{tab:roto_table_human_eval}
\end{table}

{As can be seen, texts generated by pretrained language model BART exhibits advantages in terms of coherence and fluency. On the other hand, PLMs equipped with our Structured Knowledge Hunter achieve superior information coverage and accuracy.
While coherence and fluency are important, providing accurate and comprehensive information is crucial in tasks prioritizing factual correctness, such as knowledge-grounded text generation or fact verification. 
By emphasizing information coverage and accuracy, our method ensures that the generated text is reliable and trustworthy.}


\begin{table*}[htbp]
    \caption{A hypothetical example from the Rotowire-FG dataset for an NBA game of 5 possible knowledge plans and the corresponding sentence realizations. The summarization of the performance is highlighted in Boldfaced.}
    \centering
    \small
    \setlength\tabcolsep{5.5pt}
    \begin{tabular}{l|ccccccccc}
    \hline
    \multicolumn{10}{c}{\textbf{Input Table}} \\
    \hline
    \textbf{Team} & \textbf{Name} & \textbf{City} & \textbf{Wins} &  \textbf{Losses} & \textbf{Points} & \textbf{Arena} &\textbf{1/4 Points} &\textbf{FG\_PCT} &\textbf{FG3\_PCT}\\
    \hline
    Brooklyn Nets & Net & Brooklyn & \textcolor{blue}{\textbf{37}} & \textcolor{blue}{\textbf{42}} & \textcolor{blue}{\textbf{117}} & \textcolor{blue}{\textbf{Barclays\_Center}}  & \textcolor{orange}{\textbf{31}} & \textcolor{teal}{\textbf{51}} & \textcolor{teal}{\textbf{60}} \\
    Washington Wizards & Wizard & Washington & \textcolor{blue}{\textbf{45}} & \textcolor{blue}{\textbf{34}} & \textcolor{blue}{\textbf{80}} & N/A   & \textcolor{orange}{\textbf{14}}  & \textcolor{teal}{\textbf{39}} & \textcolor{teal}{\textbf{33}} \\

    \hline
    \textbf{Player} & \textbf{Points} & \textbf{FGM} & \textbf{FGA} &  \textbf{FTM} & \textbf{FTA} & \textbf{Reb} & \textbf{FGM} & \textbf{FGA} & \textbf{Minutes} \\
    
    \hline
    Brook\_Lopez & \textcolor{violet}{\textbf{26}} & \textcolor{violet}{\textbf{12}} & \textcolor{violet}{\textbf{22}} & \textcolor{violet}{\textbf{2}} & \textcolor{violet}{\textbf{3}} & \textcolor{violet}{\textbf{9}} & & & \textcolor{violet}{\textbf{32}}  \\
    Bojan\_Bogdanovic & \textcolor{purple}{\textbf{12}} & \textcolor{purple}{\textbf{7}} & \textcolor{purple}{\textbf{12}} & 2 & 2 & & \textcolor{purple}{\textbf{6}} & \textcolor{purple}{\textbf{6}} & \textcolor{purple}{\textbf{30}}      \\
    
    \hline
    \multicolumn{10}{c}{\textbf{Knowledge Plan}} \\
    \hline
    \end{tabular}
    \begin{tabular}{p{170mm}}
        {\textbf{S1:}  {[}Nets, city, Brooklyn{]}, {[}Nets, name, Nets{]}, {[}Nets, \textcolor{blue}{\textbf{wins}}, 37{]}, {[}Nets, \textcolor{blue}{\textbf{losses}}, 42{]}, {[}Wizards, city, Washington{]}, {[}Wizards, name, Wizards{]}, {[}Wizards, \textcolor{blue}{\textbf{wins}}, 45{]}, {[}Wizards, \textcolor{blue}{\textbf{losses}}, 34{]}, {[}Nets, \textcolor{blue}{\textbf{points}}, 117{]}, {[}Wizards, \textcolor{blue}{\textbf{points}}, 80{]}, {[}Nets, \textcolor{blue}{arena}, Barclays\_Center{]}, {[}Nets, city, Brooklyn{]}{]}.}
        \\
        {\textbf{S2:} {[}Nets, name, Nets{]}, {[}Nets, \textcolor{orange}{\textbf{first\_quarter\_points}}, 31{]}, {[}Wizards, \textcolor{orange}{\textbf{first\_quarter\_points}},14{]}.}\\
        {\textbf{S3:}  {[}Nets, name, Nets{]}, {[}Nets, \textcolor{teal}{\textbf{FG\_PCT}}, 51{]}, {[}Nets, \textcolor{teal}{\textbf{FG3\_PCT}},60{]}, {[}Wizards, name, Wizards{]}, {[}Wizards, \textcolor{teal}{\textbf{FG\_PCT}}, 39{]}, {[}Wizards, \textcolor{teal}{\textbf{FG3\_PCT}},33{]}.}\\
        \textbf{S4:} {[}Brook\_Lopez, name, Brook\_Lopez{]}, {[}Brook\_Lopez, \textcolor{violet}{\textbf{points}}, 26{]}, {[}Brook\_Lopez, \textcolor{violet}{\textbf{FGM}}, 12{]}, {[}Brook\_Lopez, \textcolor{violet}{\textbf{FGA}}, 12{]}, {[}Brook\_Lopez, \textcolor{violet}{\textbf{FTM}}, 2{]}, {[}Brook\_Lopez, \textcolor{violet}{\textbf{FTA}}, 3{]}, {[}Brook\_Lopez, \textcolor{violet}{\textbf{rebound}}, 9{]}, {[}Brook\_Lopez, \textcolor{violet}{\textbf{minutes}}, 32{]}\\
        \textbf{S5:} {[}Bojan\_Bogdanovic, name, Bojan\_Bogdanovic{]}, {[}Bojan\_Bogdanovic, \textcolor{purple}{points}, 12{]}, {[}Bojan\_Bogdanovic, \textcolor{purple}{\textbf{FGM}}, 7{]}, {[}Bojan\_Bogdanovic, \textcolor{purple}{\textbf{FGA}}, 12{]}, {[}Bojan\_Bogdanovic, \textcolor{purple}{\textbf{FG3M}}, 6{]}, {[}Bojan\_Bogdanovic, \textcolor{purple}{\textbf{FG3A}}, 6{]}, {[}Bojan\_Bogdanovic, \textcolor{purple}{\textbf{FTM}}, 2{]}, {[}Bojan\_Bogdanovic, \textcolor{purple}{\textbf{FTA}}, 2{]}, {[}Bojan\_Bogdanovic, \textcolor{purple}{\textbf{minutes}}, 30{]} \\
        \hline
    \end{tabular}
    \begin{tabular}{lcccccc}
        \multicolumn{7}{c}{\textbf{Realization}} \\
    \end{tabular}    
    \begin{tabular}{p{17cm}}
    \hline
    \textbf{S1: } The Brooklyn Nets (\textcolor{blue}{\textbf{37}}-\textcolor{blue}{\textbf{42}}) defeated the Washington Wizards (\textcolor{blue}{\textbf{45}} -\textcolor{blue}{\textbf{34}} ) \textcolor{blue}{\textbf{117}}-\textcolor{blue}{\textbf{80}} on Monday at the \textcolor{blue}{\textbf{Barclays Center}} in Brooklyn.\\
    \textbf{S2: } The Nets \textbf{\textit{got off to a quick start}} in this game, outscoring the Wizards \textcolor{orange}{\textbf{31}}-\textcolor{orange}{\textbf{14}} in the first quarter alone.\\
    \textbf{S3: } The Nets \textbf{\textit{defeated}} the Wizards with a \textbf{\textit{strong shooting performance}}, led by a \textcolor{teal}{\textbf{51\%}} field goal percentage and a \textcolor{teal}{\textbf{60\%}} from beyond the arc. The Wizards \textbf{\textit{struggled to keep up}}, shooting \textcolor{teal}{\textbf{39\%}} from the field and \textcolor{teal}{\textbf{33\%}} from three-point range.\\
    \textbf{S4: } Brook\_Lopez finished with \textcolor{violet}{\textbf{26}} points ( \textcolor{violet}{\textbf{12}} - \textcolor{violet}{\textbf{22}} FG, \textcolor{violet}{\textbf{2}} - \textcolor{violet}{\textbf{3}} FT ) and \textcolor{violet}{\textbf{9}} rebounds in \textcolor{violet}{\textbf{32}} minutes.\\
    \textbf{S5: } While Bojan\_Bogdanovic finished with \textcolor{purple}{\textbf{22}} points ( \textcolor{purple}{\textbf{7}} - \textcolor{purple}{\textbf{12}} FG, \textcolor{purple}{\textbf{6}} - \textcolor{purple}{\textbf{6}} 3PT, \textcolor{purple}{\textbf{2}} - \textcolor{purple}{\textbf{2}} FT  ) in \textcolor{purple}{\textbf{30}} minutes. \\
    \hline
    \end{tabular}
    \label{tab:roto_example}
\end{table*}

\subsection{Error Analysis}

{Having established that our Structured Knowledge Hunter can improve performance in many settings, we now move on to the research question: How do the failure modes of our method compare to other methods for knowledge-enhanced text generation?}

{1.	The structured knowledge hunter may encounter difficulties in generating informative responses when the dialogue does not explicitly demand specific knowledge. This can result in an empty knowledge set being extracted, which is less satisfactory compared to utilizing the complete knowledge graph for end-to-end generation. Consequently, the dialogue response may be less informative.}

{2.	The structured knowledge hunter may face challenges in accurately extracting knowledge tuples when the context does not explicitly indicate specific entities and relationships. As a result, the model may encounter difficulties in extracting the relevant knowledge that aligns with the intended meaning of the dialogue. }

{3.	The structured knowledge hunter prioritizes factualness in text generation, it may rely heavily on structured knowledge sources to ensure the accuracy and reliability of the generated content. While this emphasis on factualness is beneficial for generating informative and trustworthy text, it can also lead to reduced diversity, especially when generating longer texts.}

\subsection{Discussion}
{Our structured knowledge hunter, with its hierarchical encoder and planner architecture, is a task-agnostic model capable of effectively handling both tables and knowledge graphs. Our model surpasses previous approaches by achieving superior knowledge selection accuracy and text generation quality while maintaining a reasonable increase in model size. In domains with incomplete, inaccurate, or poorly structured data, our hierarchical planner enhances the model's reliability as it has an additional level of assurance to select knowledge tuples from the queried entity.}

\subsection{Case Study}
\begin{figure*}[htbp]
\begin{center}
\includegraphics[width=0.95\textwidth]{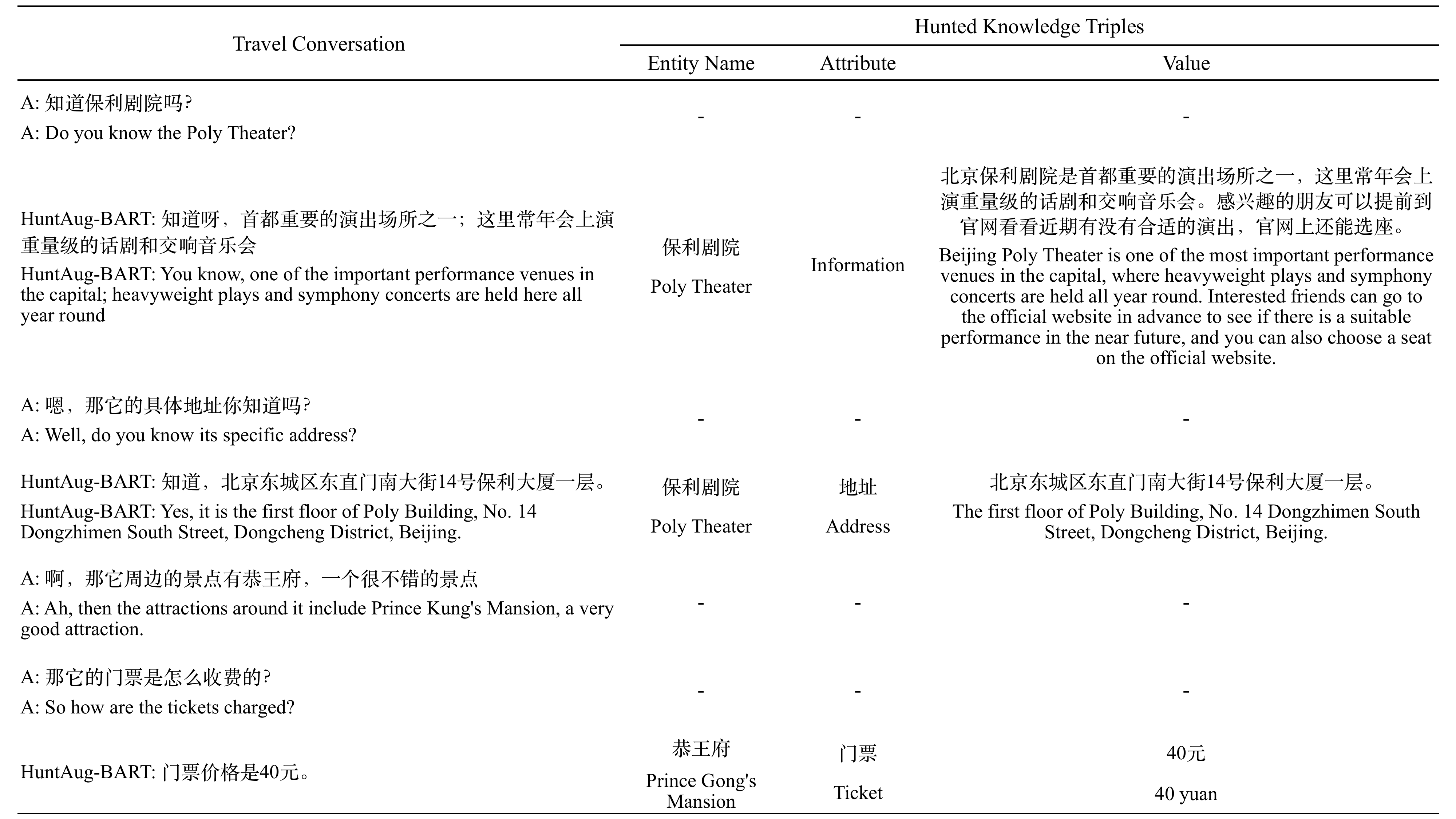}
\end{center}
\caption{A hypothetical example from the KdConv dataset and responses generated by our model HuntAug-BART for a travel conversation.  }
\label{travel_case}
\end{figure*}

Table \ref{tab:roto_example} presents a case study from the Rotowire-FG test dataset, which involves summarizing a table with redundant information. Our structured knowledge hunter first extracts representative knowledge tuples from the table to create a knowledge plan. Then, based on this plan, PLM generates the surface realization of the summary. 
Figure \ref{travel_case} presents a sample conversation from the KdConv test dataset in the travel domain between user A and HunterAug-BART. We observe that HunterAug-BART effectively selects the most relevant knowledge for user A's questions and generates responses that are both faithful and fluent.

\section{Conclusion}
In this paper, we propose a novel structured knowledge hunter that enhances interpretability in knowledge-enhanced text generation. Our method directly leverages the two-tier architecture of structured knowledge, consisting of high-level entities and low-level knowledge triples, to design our task-agnostic structured knowledge hunter. 
By combining the strong generative capabilities of language models with the faithful knowledge-hunting process, our model achieves high interpretability, enabling users to understand the model's output generation process. 
We empirically demonstrate the effectiveness of our model in both internal knowledge-enhanced table-to-text generation and external knowledge-enhanced dialogue response generation tasks. 
Our task-agnostic model outperforms state-of-the-art methods and corresponding language models, establishing new benchmarks while maintaining a high level of interpretability.

\bibliographystyle{IEEEtran}
\bibliography{custom}

@article{brown2020language,
  title={Language models are few-shot learners},
  author={Brown, Tom and Mann, Benjamin and Ryder, Nick and Subbiah, Melanie and Kaplan, Jared D and Dhariwal, Prafulla and Neelakantan, Arvind and Shyam, Pranav and Sastry, Girish and Askell, Amanda and others},
  journal={Advances in neural information processing systems},
  volume={33},
  pages={1877--1901},
  year={2020}
}

@inproceedings{wang2019revisiting,
  title={Revisiting Challenges in Data-to-Text Generation with Fact Grounding},
  author={Wang, Hongmin},
  booktitle={Proceedings of the 12th International Conference on Natural Language Generation},
  pages={311--322},
  year={2019}
}

@inproceedings{wiseman2017challenges,
  title={Challenges in Data-to-Document Generation},
  author={Wiseman, Sam and Shieber, Stuart M and Rush, Alexander M},
  booktitle={Proceedings of the 2017 Conference on Empirical Methods in Natural Language Processing},
  pages={2253--2263},
  year={2017}
}

@inproceedings{li2021improving,
  title={Improving Encoder by Auxiliary Supervision Tasks for Table-to-Text Generation},
  author={Li, Liang and Ma, Can and Yue, Yinliang and Hu, Dayong},
  booktitle={Proceedings of the 59th Annual Meeting of the Association for Computational Linguistics and the 11th International Joint Conference on Natural Language Processing (Volume 1: Long Papers)},
  pages={5979--5989},
  year={2021}
}

@inproceedings{puduppully-etal-2019-data,
    title = "Data-to-text Generation with Entity Modeling",
    author = "Puduppully, Ratish  and
      Dong, Li  and
      Lapata, Mirella",
    booktitle = "Proceedings of the 57th Annual Meeting of the Association for Computational Linguistics",
    month = jul,
    year = "2019",
    address = "Florence, Italy",
    publisher = "Association for Computational Linguistics",
    url = "https://aclanthology.org/P19-1195",
    doi = "10.18653/v1/P19-1195",
    pages = "2023--2035",
}

@inproceedings{lebret2016neural,
  title={Neural Text Generation from Structured Data with Application to the Biography Domain},
  author={Lebret, R{\'e}mi and Grangier, David and Auli, Michael},
  booktitle={Proceedings of the 2016 Conference on Empirical Methods in Natural Language Processing},
  pages={1203--1213},
  year={2016}
}

@inproceedings{puduppully2019data,
  title={Data-to-text generation with content selection and planning},
  author={Puduppully, Ratish and Dong, Li and Lapata, Mirella},
  booktitle={Proceedings of the AAAI conference on artificial intelligence},
  volume={33},
  number={01},
  pages={6908--6915},
  year={2019}
}

@inproceedings{zhou2020kdconv,
  title={KdConv: A Chinese Multi-domain Dialogue Dataset Towards Multi-turn Knowledge-driven Conversation},
  author={Zhou, Hao and Zheng, Chujie and Huang, Kaili and Huang, Minlie and Zhu, Xiaoyan},
  booktitle={Proceedings of the 58th Annual Meeting of the Association for Computational Linguistics},
  pages={7098--7108},
  year={2020}
}

@inproceedings{lewis2020bart,
  title={BART: Denoising Sequence-to-Sequence Pre-training for Natural Language Generation, Translation, and Comprehension},
  author={Lewis, Mike and Liu, Yinhan and Goyal, Naman and Ghazvininejad, Marjan and Mohamed, Abdelrahman and Levy, Omer and Stoyanov, Veselin and Zettlemoyer, Luke},
  booktitle={Proceedings of the 58th Annual Meeting of the Association for Computational Linguistics},
  pages={7871--7880},
  year={2020}
}

@article{raffel2020exploring,
  title={Exploring the limits of transfer learning with a unified text-to-text transformer.},
  author={Raffel, Colin and Shazeer, Noam and Roberts, Adam and Lee, Katherine and Narang, Sharan and Matena, Michael and Zhou, Yanqi and Li, Wei and Liu, Peter J and others},
  journal={J. Mach. Learn. Res.},
  volume={21},
  number={140},
  pages={1--67},
  year={2020}
}

@inproceedings{gong2019table,
  title={Table-to-Text Generation with Effective Hierarchical Encoder on Three Dimensions (Row, Column and Time)},
  author={Gong, Heng and Feng, Xiaocheng and Qin, Bing and Liu, Ting},
  booktitle={Proceedings of the 2019 Conference on Empirical Methods in Natural Language Processing and the 9th International Joint Conference on Natural Language Processing (EMNLP-IJCNLP)},
  pages={3143--3152},
  year={2019}
}

@article{puduppully2021data,
  title={Data-to-text generation with macro planning},
  author={Puduppully, Ratish and Lapata, Mirella},
  journal={Transactions of the Association for Computational Linguistics},
  volume={9},
  pages={510--527},
  year={2021},
  publisher={MIT Press}
}

@inproceedings{kenton2019bert,
  title={BERT: Pre-training of Deep Bidirectional Transformers for Language Understanding},
  author={Kenton, Jacob Devlin Ming-Wei Chang and Toutanova, Lee Kristina},
  booktitle={Proceedings of NAACL-HLT},
  pages={4171--4186},
  year={2019}
}

@inproceedings{serban2016building,
  title={Building end-to-end dialogue systems using generative hierarchical neural network models},
  author={Serban, Iulian and Sordoni, Alessandro and Bengio, Yoshua and Courville, Aaron and Pineau, Joelle},
  booktitle={Proceedings of the AAAI Conference on Artificial Intelligence},
  volume={30},
  number={1},
  year={2016}
}

@inproceedings{DBLP:conf/aaai/SerbanSBCP16,
  author       = {Iulian Vlad Serban and
                  Alessandro Sordoni and
                  Yoshua Bengio and
                  Aaron C. Courville and
                  Joelle Pineau},
  editor       = {Dale Schuurmans and
                  Michael P. Wellman},
  title        = {Building End-To-End Dialogue Systems Using Generative Hierarchical
                  Neural Network Models},
  booktitle    = {Proceedings of the Thirtieth {AAAI} Conference on Artificial Intelligence,
                  February 12-17, 2016, Phoenix, Arizona, {USA}},
  pages        = {3776--3784},
  publisher    = {{AAAI} Press},
  year         = {2016},
  url          = {https://doi.org/10.1609/aaai.v30i1.9883},
  doi          = {10.1609/AAAI.V30I1.9883},
  bibsource    = {dblp computer science bibliography, https://dblp.org}
}

@inproceedings{DBLP:conf/iclr/LoshchilovH19,
  author       = {Ilya Loshchilov and
                  Frank Hutter},
  title        = {Decoupled Weight Decay Regularization},
  booktitle    = {7th International Conference on Learning Representations, {ICLR} 2019,
                  New Orleans, LA, USA, May 6-9, 2019},
  publisher    = {OpenReview.net},
  year         = {2019},
  url          = {https://openreview.net/forum?id=Bkg6RiCqY7},
  bibsource    = {dblp computer science bibliography, https://dblp.org}
}

@online{sun2012jieba,
  author = {Sun, Junyi},
  title = {Jieba chinese word segmentation tool},
  year = 2012,
  url = {https://github.com/fxsjy/jieba},
}

@article{shao2021cpt,
  title={CPT: A Pre-Trained Unbalanced Transformer for Both Chinese Language Understanding and Generation}, 
  author={Yunfan Shao and Zhichao Geng and Yitao Liu and Junqi Dai and Fei Yang and Li Zhe and Hujun Bao and Xipeng Qiu},
  journal={arXiv preprint arXiv:2109.05729},
  year={2021}
}

@inproceedings{liu2021kg,
  title={Kg-bart: Knowledge graph-augmented bart for generative commonsense reasoning},
  author={Liu, Ye and Wan, Yao and He, Lifang and Peng, Hao and Philip, S Yu},
  booktitle={Proceedings of the AAAI Conference on Artificial Intelligence},
  volume={35},
  number={7},
  pages={6418--6425},
  year={2021}
}

@inproceedings{liu2020k,
  title={K-bert: Enabling language representation with knowledge graph},
  author={Liu, Weijie and Zhou, Peng and Zhao, Zhe and Wang, Zhiruo and Ju, Qi and Deng, Haotang and Wang, Ping},
  booktitle={Proceedings of the AAAI Conference on Artificial Intelligence},
  volume={34},
  number={03},
  pages={2901--2908},
  year={2020}
}

@article{sun2019ernie,
  title={Ernie: Enhanced representation through knowledge integration},
  author={Sun, Yu and Wang, Shuohuan and Li, Yukun and Feng, Shikun and Chen, Xuyi and Zhang, Han and Tian, Xin and Zhu, Danxiang and Tian, Hao and Wu, Hua},
  journal={arXiv preprint arXiv:1904.09223},
  year={2019}
}

@inproceedings{
zhou2021pretraining,
title={Pre-training Text-to-Text Transformers for Concept-centric Common Sense},
author={Wangchunshu Zhou and Dong-Ho Lee and Ravi Kiran Selvam and Seyeon Lee and Xiang Ren},
booktitle={International Conference on Learning Representations},
year={2021},
url={https://openreview.net/forum?id=3k20LAiHYL2}
}

@article{yu2020jaket,
  title={Jaket: Joint pre-training of knowledge graph and language understanding},
  author={Yu, Donghan and Zhu, Chenguang and Yang, Yiming and Zeng, Michael}
}

@article{guan2020knowledge,
  title={A Knowledge-Enhanced Pretraining Model for Commonsense Story Generation},
  author={Guan, Jian and Huang, Fei and Zhao, Zhihao and Zhu, Xiaoyan and Huang, Minlie},
  journal={Transactions of the Association for Computational Linguistics},
  volume={8},
  pages={93--108},
  year={2020}
}

@article{xu2018graph2seq,
  title={Graph2Seq: Graph to Sequence Learning with Attention-based Neural Networks},
  author={Xu, Kun and Wu, Lingfei and Wang, Zhiguo and Feng, Yansong and Witbrock, Michael and Sheinin, Vadim},
  journal={arXiv e-prints},
  pages={arXiv--1804},
  year={2018}
}

@inproceedings{xie2022unifiedskg,
  title={UnifiedSKG: Unifying and Multi-Tasking Structured Knowledge Grounding with Text-to-Text Language Models},
  author={Xie, Tianbao and Wu, Chen Henry and Shi, Peng and Zhong, Ruiqi and Scholak, Torsten and Yasunaga, Michihiro and Wu, Chien-Sheng and Zhong, Ming and Yin, Pengcheng and Wang, Sida I and others},
  booktitle={Proceedings of the 2022 Conference on Empirical Methods in Natural Language Processing},
  pages={602--631},
  year={2022}
}

@inproceedings{yasunaga2021qa,
  title={QA-GNN: Reasoning with Language Models and Knowledge Graphs for Question Answering},
  author={Yasunaga, Michihiro and Ren, Hongyu and Bosselut, Antoine and Liang, Percy and Leskovec, Jure},
  booktitle={Proceedings of the 2021 Conference of the North American Chapter of the Association for Computational Linguistics: Human Language Technologies},
  pages={535--546},
  year={2021}
}

@inproceedings{moon2019opendialkg,
  title={Opendialkg: Explainable conversational reasoning with attention-based walks over knowledge graphs},
  author={Moon, Seungwhan and Shah, Pararth and Kumar, Anuj and Subba, Rajen},
  booktitle={Proceedings of the 57th Annual Meeting of the Association for Computational Linguistics},
  pages={845--854},
  year={2019}
}

@inproceedings{borgeaud2022improving,
  title={Improving language models by retrieving from trillions of tokens},
  author={Borgeaud, Sebastian and Mensch, Arthur and Hoffmann, Jordan and Cai, Trevor and Rutherford, Eliza and Millican, Katie and Van Den Driessche, George Bm and Lespiau, Jean-Baptiste and Damoc, Bogdan and Clark, Aidan and others},
  booktitle={International conference on machine learning},
  pages={2206--2240},
  year={2022},
  organization={PMLR}
}

@article{wang2019superglue,
  title={Superglue: A stickier benchmark for general-purpose language understanding systems},
  author={Wang, Alex and Pruksachatkun, Yada and Nangia, Nikita and Singh, Amanpreet and Michael, Julian and Hill, Felix and Levy, Omer and Bowman, Samuel},
  journal={Advances in neural information processing systems},
  volume={32},
  year={2019}
}

@article{devlin2018bert,
  title={Bert: Pre-training of deep bidirectional transformers for language understanding},
  author={Devlin, Jacob and Chang, Ming-Wei and Lee, Kenton and Toutanova, Kristina},
  journal={arXiv preprint arXiv:1810.04805},
  year={2018}
}

@article{radford2019language,
  title={Language models are unsupervised multitask learners},
  author={Radford, Alec and Wu, Jeffrey and Child, Rewon and Luan, David and Amodei, Dario and Sutskever, Ilya and others},
  journal={OpenAI blog},
  volume={1},
  number={8},
  pages={9},
  year={2019}
}

@article{nogueira2019passage,
  title={Passage Re-ranking with BERT},
  author={Nogueira, Rodrigo and Cho, Kyunghyun},
  journal={arXiv preprint arXiv:1901.04085},
  year={2019}
}

@inproceedings{karpukhin-etal-2020-dense,
    title = "Dense Passage Retrieval for Open-Domain Question Answering",
    author = "Karpukhin, Vladimir  and
      Oguz, Barlas  and
      Min, Sewon  and
      Lewis, Patrick  and
      Wu, Ledell  and
      Edunov, Sergey  and
      Chen, Danqi  and
      Yih, Wen-tau",
    editor = "Webber, Bonnie  and
      Cohn, Trevor  and
      He, Yulan  and
      Liu, Yang",
    booktitle = "Proceedings of the 2020 Conference on Empirical Methods in Natural Language Processing (EMNLP)",
    month = nov,
    year = "2020",
    address = "Online",
    publisher = "Association for Computational Linguistics",
    url = "https://aclanthology.org/2020.emnlp-main.550",
    doi = "10.18653/v1/2020.emnlp-main.550",
    pages = "6769--6781",
}

@inproceedings{guu2020retrieval,
  title={Retrieval augmented language model pre-training},
  author={Guu, Kelvin and Lee, Kenton and Tung, Zora and Pasupat, Panupong and Chang, Mingwei},
  booktitle={International conference on machine learning},
  pages={3929--3938},
  year={2020},
  organization={PMLR}
}

@article{yin2022survey,
  title={A survey of knowledge-intensive nlp with pre-trained language models},
  author={Yin, Da and Dong, Li and Cheng, Hao and Liu, Xiaodong and Chang, Kai-Wei and Wei, Furu and Gao, Jianfeng},
  journal={arXiv preprint arXiv:2202.08772},
  year={2022}
}

@article{thoppilan2022lamda,
  title={Lamda: Language models for dialog applications},
  author={Thoppilan, Romal and De Freitas, Daniel and Hall, Jamie and Shazeer, Noam and Kulshreshtha, Apoorv and Cheng, Heng-Tze and Jin, Alicia and Bos, Taylor and Baker, Leslie and Du, Yu and others},
  journal={arXiv preprint arXiv:2201.08239},
  year={2022}
}

@article{chowdhery2022palm,
  title={Palm: Scaling language modeling with pathways},
  author={Chowdhery, Aakanksha and Narang, Sharan and Devlin, Jacob and Bosma, Maarten and Mishra, Gaurav and Roberts, Adam and Barham, Paul and Chung, Hyung Won and Sutton, Charles and Gehrmann, Sebastian and others},
  journal={arXiv preprint arXiv:2204.02311},
  year={2022}
}

@inproceedings{sun2022black,
  title={Black-box tuning for language-model-as-a-service},
  author={Sun, Tianxiang and Shao, Yunfan and Qian, Hong and Huang, Xuanjing and Qiu, Xipeng},
  booktitle={International Conference on Machine Learning},
  pages={20841--20855},
  year={2022},
  organization={PMLR}
}

@article{diao2022black,
  title={Black-Box Prompt Learning for Pre-trained Language Models},
  author={Diao, Shizhe and Huang, Zhichao and Xu, Ruijia and Li, Xuechun and Yong, LIN and Zhou, Xiao and Zhang, Tong},
  journal={Transactions on Machine Learning Research},
  year={2022}
}

@article{mialon2023augmented,
  title={Augmented language models: a survey},
  author={Mialon, Gr{\'e}goire and Dess{\`\i}, Roberto and Lomeli, Maria and Nalmpantis, Christoforos and Pasunuru, Ram and Raileanu, Roberta and Rozi{\`e}re, Baptiste and Schick, Timo and Dwivedi-Yu, Jane and Celikyilmaz, Asli and others},
  journal={arXiv preprint arXiv:2302.07842},
  year={2023}
}

@article{creswell2022selection,
  title={Selection-inference: Exploiting large language models for interpretable logical reasoning},
  author={Creswell, Antonia and Shanahan, Murray and Higgins, Irina},
  journal={arXiv preprint arXiv:2205.09712},
  year={2022}
}

@article{liu2023pre,
  title={Pre-train, prompt, and predict: A systematic survey of prompting methods in natural language processing},
  author={Liu, Pengfei and Yuan, Weizhe and Fu, Jinlan and Jiang, Zhengbao and Hayashi, Hiroaki and Neubig, Graham},
  journal={ACM Computing Surveys},
  volume={55},
  number={9},
  pages={1--35},
  year={2023},
  publisher={ACM New York, NY}
}

@inproceedings{ni2022hitkg,
  title={Hitkg: Towards goal-oriented conversations via multi-hierarchy learning},
  author={Ni, Jinjie and Pandelea, Vlad and Young, Tom and Zhou, Haicang and Cambria, Erik},
  booktitle={Proceedings of the AAAI conference on artificial intelligence},
  volume={36},
  number={10},
  pages={11112--11120},
  year={2022}
}

@article{zhong2023knowledge,
  title={Knowledge graph augmented network towards multiview representation learning for aspect-based sentiment analysis},
  author={Zhong, Qihuang and Ding, Liang and Liu, Juhua and Du, Bo and Jin, Hua and Tao, Dacheng},
  journal={IEEE Transactions on Knowledge and Data Engineering},
  year={2023},
  publisher={IEEE}
}

@article{menick2022teaching,
  title={Teaching language models to support answers with verified quotes},
  author={Menick, Jacob and Trebacz, Maja and Mikulik, Vladimir and Aslanides, John and Song, Francis and Chadwick, Martin and Glaese, Mia and Young, Susannah and Campbell-Gillingham, Lucy and Irving, Geoffrey and others},
  journal={arXiv preprint arXiv:2203.11147},
  year={2022}
}

@inproceedings{liu2022rainier,
  title={Rainier: Reinforced Knowledge Introspector for Commonsense Question Answering},
  author={Liu, Jiacheng and Hallinan, Skyler and Lu, Ximing and He, Pengfei and Welleck, Sean and Hajishirzi, Hannaneh and Choi, Yejin},
  booktitle={Proceedings of the 2022 Conference on Empirical Methods in Natural Language Processing},
  pages={8938--8958},
  year={2022}
}

@inproceedings{zhang2022subgraph,
  title={Subgraph Retrieval Enhanced Model for Multi-hop Knowledge Base Question Answering},
  author={Zhang, Jing and Zhang, Xiaokang and Yu, Jifan and Tang, Jian and Tang, Jie and Li, Cuiping and Chen, Hong},
  booktitle={Proceedings of the 60th Annual Meeting of the Association for Computational Linguistics (Volume 1: Long Papers)},
  pages={5773--5784},
  year={2022}
}

@article{puduppully2022data,
  title={Data-to-text generation with variational sequential planning},
  author={Puduppully, Ratish and Fu, Yao and Lapata, Mirella},
  journal={Transactions of the Association for Computational Linguistics},
  volume={10},
  pages={697--715},
  year={2022},
  publisher={MIT Press}
}

@inproceedings{li2023plan,
  title={Plan-then-Seam: Towards Efficient Table-to-Text Generation},
  author={Li, Liang and Geng, Ruiying and Fang, Chengyang and Li, Bing and Ma, Can and Li, Binhua and Li, Yongbin},
  booktitle={Findings of the Association for Computational Linguistics: EACL 2023},
  pages={205--219},
  year={2023}
}

@inproceedings{zhao-etal-2022-standard,
    title = "There Is No Standard Answer: Knowledge-Grounded Dialogue Generation with Adversarial Activated Multi-Reference Learning",
    author = "Zhao, Xueliang  and
      Fu, Tingchen  and
      Tao, Chongyang  and
      Yan, Rui",
    editor = "Goldberg, Yoav  and
      Kozareva, Zornitsa  and
      Zhang, Yue",
    booktitle = "Proceedings of the 2022 Conference on Empirical Methods in Natural Language Processing",
    month = dec,
    year = "2022",
    address = "Abu Dhabi, United Arab Emirates",
    publisher = "Association for Computational Linguistics",
    url = "https://aclanthology.org/2022.emnlp-main.123",
    doi = "10.18653/v1/2022.emnlp-main.123",
    pages = "1878--1891",
}

@inproceedings{shuster-etal-2022-language,
    title = "Language Models that Seek for Knowledge: Modular Search {\&} Generation for Dialogue and Prompt Completion",
    author = "Shuster, Kurt  and
      Komeili, Mojtaba  and
      Adolphs, Leonard  and
      Roller, Stephen  and
      Szlam, Arthur  and
      Weston, Jason",
    editor = "Goldberg, Yoav  and
      Kozareva, Zornitsa  and
      Zhang, Yue",
    booktitle = "Findings of the Association for Computational Linguistics: EMNLP 2022",
    month = dec,
    year = "2022",
    address = "Abu Dhabi, United Arab Emirates",
    publisher = "Association for Computational Linguistics",
    url = "https://aclanthology.org/2022.findings-emnlp.27",
    doi = "10.18653/v1/2022.findings-emnlp.27",
    pages = "373--393",
}

@inproceedings{wang2022rt,
  title={RT-KGD: Relation Transition Aware Knowledge-Grounded Dialogue Generation},
  author={Wang, Kexin and Li, Zhixu and Wang, Jiaan and Qu, Jianfeng and He, Ying and Liu, An and Zhao, Lei},
  booktitle={The Semantic Web--ISWC 2022: 21st International Semantic Web Conference, Virtual Event, October 23--27, 2022, Proceedings},
  pages={319--335},
  year={2022},
  organization={Springer}
}

@inproceedings{sarkar2022kg,
  title={KG-CRuSE: Recurrent Walks over Knowledge Graph for Explainable Conversation Reasoning using Semantic Embeddings},
  author={Sarkar, Rajdeep and Arcan, Mihael and McCrae, John Philip},
  booktitle={Proceedings of the 4th Workshop on NLP for Conversational AI},
  pages={98--107},
  year={2022}
}

@inproceedings{
kang2022knowledgeconsistent,
title={Knowledge-Consistent Dialogue Generation with Knowledge Graphs},
author={Minki Kang and Jin Myung Kwak and Jinheon Baek and Sung Ju Hwang},
booktitle={ICML 2022 Workshop on Knowledge Retrieval and Language Models},
year={2022},
url={https://openreview.net/forum?id=McHtKDi5h9}
}

@inproceedings{tuan-etal-2022-towards,
    title = "Towards Large-Scale Interpretable Knowledge Graph Reasoning for Dialogue Systems",
    author = "Tuan, Yi-Lin  and
      Beygi, Sajjad  and
      Fazel-Zarandi, Maryam  and
      Gao, Qiaozi  and
      Cervone, Alessandra  and
      Wang, William Yang",
    editor = "Muresan, Smaranda  and
      Nakov, Preslav  and
      Villavicencio, Aline",
    booktitle = "Findings of the Association for Computational Linguistics: ACL 2022",
    month = may,
    year = "2022",
    address = "Dublin, Ireland",
    publisher = "Association for Computational Linguistics",
    url = "https://aclanthology.org/2022.findings-acl.33",
    doi = "10.18653/v1/2022.findings-acl.33",
    pages = "383--395",
}

@inproceedings{yu2022jaket,
  title={Jaket: Joint pre-training of knowledge graph and language understanding},
  author={Yu, Donghan and Zhu, Chenguang and Yang, Yiming and Zeng, Michael},
  booktitle={Proceedings of the AAAI Conference on Artificial Intelligence},
  volume={36},
  number={10},
  pages={11630--11638},
  year={2022}
}

@inproceedings{yang-etal-2022-tableformer,
    title = "{T}able{F}ormer: Robust Transformer Modeling for Table-Text Encoding",
    author = "Yang, Jingfeng  and
      Gupta, Aditya  and
      Upadhyay, Shyam  and
      He, Luheng  and
      Goel, Rahul  and
      Paul, Shachi",
    editor = "Muresan, Smaranda  and
      Nakov, Preslav  and
      Villavicencio, Aline",
    booktitle = "Proceedings of the 60th Annual Meeting of the Association for Computational Linguistics (Volume 1: Long Papers)",
    month = may,
    year = "2022",
    address = "Dublin, Ireland",
    publisher = "Association for Computational Linguistics",
    url = "https://aclanthology.org/2022.acl-long.40",
    doi = "10.18653/v1/2022.acl-long.40",
    pages = "528--537",
}

@inproceedings{NIPS2000_728f206c,
 author = {Bengio, Yoshua and Ducharme, R\'{e}jean and Vincent, Pascal},
 booktitle = {Advances in Neural Information Processing Systems},
 editor = {T. Leen and T. Dietterich and V. Tresp},
 pages = {},
 publisher = {MIT Press},
 title = {A Neural Probabilistic Language Model},
 url = {https://proceedings.neurips.cc/paper_files/paper/2000/file/728f206c2a01bf572b5940d7d9a8fa4c-Paper.pdf},
 volume = {13},
 year = {2000}
}

@article{sutskever2014sequence,
  title={Sequence to sequence learning with neural networks},
  author={Sutskever, Ilya and Vinyals, Oriol and Le, Quoc V},
  journal={Advances in neural information processing systems},
  volume={27},
  year={2014}
}

@inproceedings{aly2023qa,
  title={QA-NatVer: Question Answering for Natural Logic-based Fact Verification},
  author={Aly, Rami and Strong, Marek and Vlachos, Andreas},
  booktitle={Proceedings of the 2023 Conference on Empirical Methods in Natural Language Processing},
  pages={8376--8391},
  year={2023}
}

@article{yuan2023zero,
  title={Zero-Shot Fact-Checking with Semantic Triples and Knowledge Graphs},
  author={Yuan, Zhangdie and Vlachos, Andreas},
  journal={arXiv preprint arXiv:2312.11785},
  year={2023}
}

@inproceedings{DBLP:conf/iclr/ZhangKWWA20,
  author       = {Tianyi Zhang and
                  Varsha Kishore and
                  Felix Wu and
                  Kilian Q. Weinberger and
                  Yoav Artzi},
  title        = {BERTScore: Evaluating Text Generation with {BERT}},
  booktitle    = {8th International Conference on Learning Representations, {ICLR} 2020,
                  Addis Ababa, Ethiopia, April 26-30, 2020},
  publisher    = {OpenReview.net},
  year         = {2020},
  url          = {https://openreview.net/forum?id=SkeHuCVFDr},
  bibsource    = {dblp computer science bibliography, https://dblp.org}
}

@InProceedings{	  bengio2000neural,
  title		= {{A Neural Probabilistic Language Model}},
  author	= {Bengio, Yoshua and Ducharme, R{\'e}jean and Vincent, Pascal},
  booktitle	= nips,
  volume	= {13},
  year	= {2000}
}

\vfill

\end{document}